\documentclass[runningheads]{llncs}
\usepackage{graphicx}
\usepackage{subfigure}

\usepackage{tikz}
\usepackage{comment}
\usepackage{amsmath,amssymb} 
\usepackage{color}

\usepackage{amsmath}
\usepackage{amssymb}
\usepackage{booktabs}
\usepackage{bm}
\usepackage{multirow}
\usepackage{appendix}

\usepackage{makecell}

\usepackage[accsupp]{axessibility}  

\newcommand{\ie}{\textit{i}.\textit{e}. }
\newcommand{\eg}{\textit{e}.\textit{g}. }
\newcommand{\wrt}{\textit{w}.\textit{r}.\textit{t}. }

\begin{document}
\pagestyle{headings}
\mainmatter
\def\ECCVSubNumber{6660}  

\title{Entry-Flipped Transformer for Inference and Prediction of Participant Behavior} 

\titlerunning{EF-Transformer for Inference and Prediction of Participant Behavior}
%
\author{Bo Hu\inst{1,2} \and
Tat-Jen Cham\inst{1,2}}
\authorrunning{B. Hu and T. Cham}
%
\institute{Singtel Cognitive and Artificial Intelligence Lab (SCALE@NTU), Singapore \and School of Computer Science and Engineering, Nanyang Technological University, Singapore \\
\email{\{hubo,astjcham\}@ntu.edu.sg}}
\maketitle

\begin{abstract}
Some group activities, such as team sports and choreographed dances, involve closely coupled interaction between participants. Here we investigate the tasks of inferring and predicting participant behavior, in terms of motion paths and actions, under such conditions. We narrow the problem to that of estimating how a set target participants react to the behavior of other observed participants. Our key idea is to model the spatio-temporal relations among participants in a manner that is robust to error accumulation during frame-wise inference and prediction. We propose a novel Entry-Flipped Transformer (EF-Transformer), which models the relations of participants by attention mechanisms on both spatial and temporal domains. Unlike typical transformers, we tackle the problem of error accumulation by flipping the order of query, key, and value entries, to increase the importance and fidelity of observed features in the current frame. Comparative experiments show that our EF-Transformer achieves the best performance on a newly-collected tennis doubles dataset, a Ceilidh dance dataset, and two pedestrian datasets. Furthermore, it is also demonstrated that our EF-Transformer is better at limiting accumulated errors and recovering from wrong estimations. 
\end{abstract}

\section{Introduction}\label{sec:intro}
The development of computer vision with machine learning has led to extensive progress in understanding human behavior, such as human action recognition and temporal action detection. Although state-of-the-art algorithms have shown promise, a majority of methods have been focused only on individuals without explicitly handling interaction between people.
However, human behavior can span a wide range of interaction coupling, from the independence of strangers passing each other, to highly coordinated activities such as in group sports and choreographed dances.

\begin{figure}[t]
    \centering
    \includegraphics[width=0.65\linewidth]{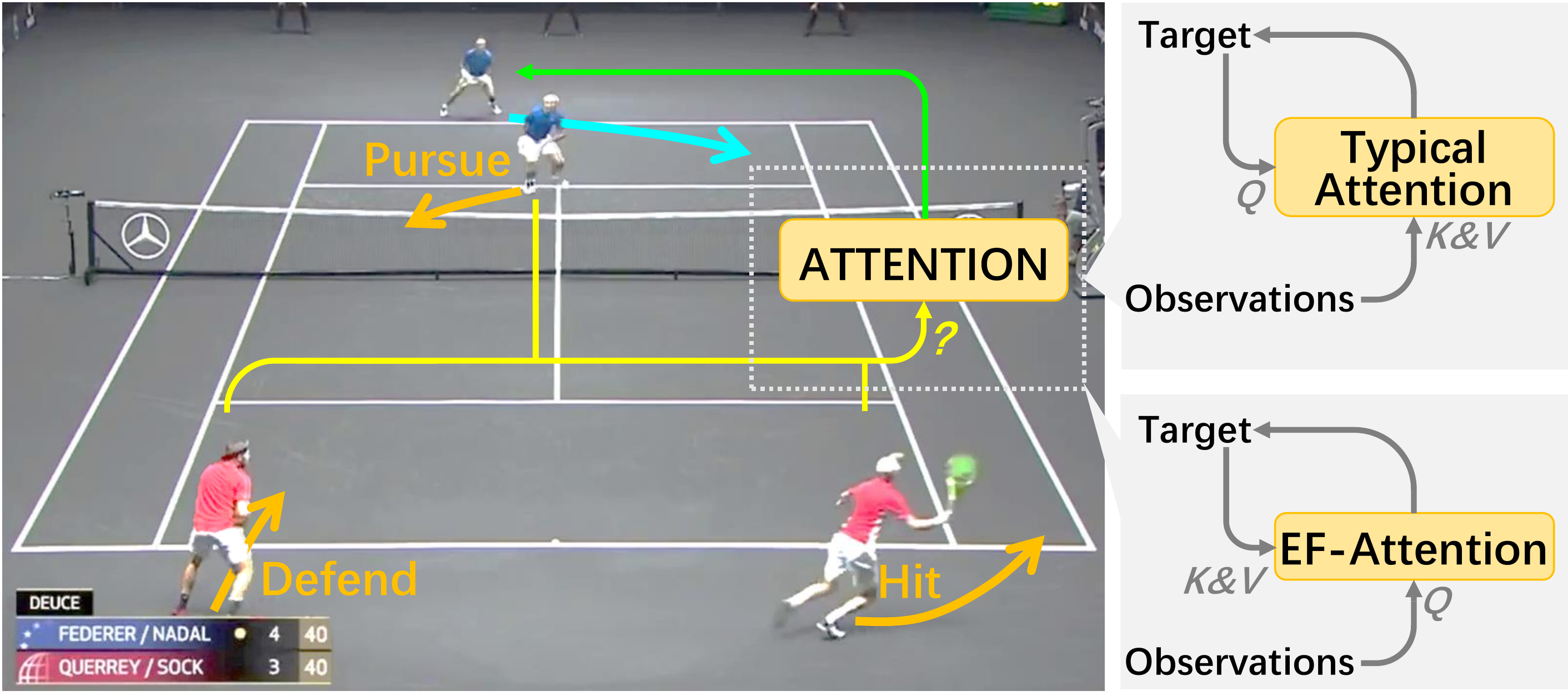}
    \caption{This paper focuses on participants behavior prediction and inference, where the behavior of target participant from a group activity is estimated with observation of other participants. Entry-Flipping (EF) mechanism is proposed for attention function to obtain accurate prediction and inference by flipping the query, key, and value entries. 
    }
    \label{fig_one}
\end{figure}

The behavior of a person can be treated as a combination of self intention and social interaction, where the latter is more crucial in group activities. Current group-related computer vision works do not focus much on scenarios with heavy social interaction among participants. For example, in pedestrian trajectory prediction \cite{alahi2016social,zhang2019sr}, the behavior of a pedestrian is based more on self intention than social interaction, with the latter cursorily for avoiding collisions.

To further explore the model of social interactions in group activities, we consider the tasks of inferring and predicting the behavior of some participants as they react to other participants. In these tasks, we hypothesize that the behavior of participants of a group activity are less dependent on self intentions, and instead dominated by how other participants behave. To formalize the problem, we consider a group as split into two sets of observed and target participants. For target participants, we assume that no data is provided beyond some initial states --- the objective is thus to infer their behavior based \emph{only} on the continuing data received from observed participants (see fig.\ \ref{fig_one}).
We believe that this modeling of reactive human behavior in closely coupled activities such as team sports, will eventually lead to enabling more realistic agent behavior models, \eg for simulation in games or sports training.

The task of inferring or predicting participant behavior is a frame-wise sequence estimation problem. There are many existing models focused on sequence estimation, such as Recurrent Neural Networks (RNN) based methods \cite{sutskever2014sequence,xu2018encoding,liang2019peeking} and attention-based methods \cite{vaswani2017attention,yu2020spatio}. However, these methods face the problem of error accumulation,
as the recurrence involves using the output estimation from the previous step as the input in the next step. While this leads to temporally smooth predictions, small errors at each step accumulate over time, leading to large final errors. Taking a typical transformer \cite{vaswani2017attention} as an example, the cross-attention in the decoder auto-regressively uses the previous estimate as query input. As the query is the base of an attention function, errors in subsequent queries will often grow, even if the key and value entries are accurate. This may not be a concern for \eg open-ended text generation, but becomes an issue for our tasks that prioritize accurate current estimates over temporal consistency.

In this paper, we propose the Entry-Flipped Transformer (EF-Transformer), a novel framework for the inference and prediction of participant behavior. Two key properties needed are: i) good relation modeling, ii) limiting the error accumulation. To model spatio-temporal relations among all participants in different frames, we adopt a transformer-based structure with multiple layers of encoders and decoders. In every encoder, separate attentions are used for the spatial domain, \ie involving different participants, and the temporal domain, \ie across different frames. Each decoder contains spatio-temporal self-attention and also cross-attention to relate features of observed and target participants. To limit accumulated errors during frame-wise inference and prediction, an entry-flipped design is introduced to the cross-attention in decoders, to focus more on correctness of output than smoothness. In our method, the query, key, and value entries of decoders are flipped \wrt the typical order. As accurate information of observed participants is sent to query entry of the attention function at each step, error accumulation can be effectively suppressed. 

The main contributions of this paper are as follows:
\begin{itemize}
    \item We articulate the key considerations needed for inferring and predicting participant behavior in group activities that involve highly coupled interactions.
    \item A novel EF-Transformer framework is proposed for this task, where query, key, value entries are flipped in cross-attention of decoders.
    \item Our method achieved SOTA performance on a tennis doubles dataset and a Ceilidh dance dataset that involve highly coupled interactions, and also outperformed other methods on looser coupled pedestrian datasets.
    \item We show our method is more robust at limiting accumulated errors and recovering from spike errors.
\end{itemize}

\section{Related Work}\label{sec:literature}
\noindent\textbf{Relation Modeling.} Participant behavior prediction involve several modules, with a core of spatio-temporal relation modeling.
Probabilistic graphical models have been used to model relations, \eg Dynamic Bayesian Networks (DBN) \cite{zeng2010knowledge}, Conditional Random Fields (CRF) \cite{amer2014hirf}, but these models heavily relied on feature engineering. 
With deep learning, models can directly learn the relations and find good features simultaneously. Convolutional Neural Networks (CNN) are widely employed to extract features from images and videos, while deeper layers of a CNN can be viewed as relation modeling since they summarize features from a larger image region \cite{lea2017temporal,qiu2017learning,carreira2017quo,azar2019convolutional}. 
Graph Convolution Networks (GCN) \cite{yan2018spatial,zeng2019graph} are used to learn the relation among features without a fixed grid format. However, convolutions usually have limited receptive fields, and are enlarged only through many layers. RNNs, such as LSTM,
 have been used to model temporal relation in sequences \cite{xu2019temporal,aliakbarian2017encouraging}. Different from CNNs processing all entries in one go, RNNs are applied iteratively over time.
Attention mechanisms were popularized by the Transformer \cite{vaswani2017attention} and became 
adopted for both spatial and temporal relation modeling \cite{zhu2021deformable,carion2020end,wang2018non}.
Attention facilitates summarization for different types of input, leading to better generalization, which  can be built upon backbone networks \cite{girdhar2019video,hu2018relation,li2020spatio}, or in feature learning \cite{dosovitskiy2020image}. However, the computational cost of attention is
large, thus many methods \cite{zhao2019bayesian,velivckovic2018graph,yuan2017temporal} are hybrids involving a combination of CNN, RNN, and attention to balance efficiency and effectiveness.

\noindent\textbf{Group Relevant Tasks.}
Group activities typically involve significant behavorial relations among group members.
Group activity recognition aims to estimate video-level activity labels. In \cite{bagautdinov2016social,ibrahim2016hierarchical,wu2019learning,ibrahim2018hierarchical} RNN was used to model temporal relation of each person and pooled all persons together for recognition. Cross inference block has been proposed in HiGCIN \cite{yan2020higcin} to capture co-occurrence spatiotemporal dependencies. In the actor transformer \cite{gavrilyuk2020actor}, the transformer encodes all actors after actor-level features are extracted. 
These frameworks are impressive but unsuitable for our proposed tasks, as they are not designed for frame-level estimation. Another related task is pedestrian trajectory prediction \cite{Lerner2007crowd,pellegrini2009you,vemula2018social,xu2018encoding}. The goal is to predict moving trajectories of all pedestrians in future frames with observation of a few past frames, where interaction among pedestrians is the important cue. RNN \cite{alahi2016social,becker2018red}, graph-based technique \cite{yu2020spatio}, and attention mechanism \cite{fernando2018soft+,sadeghian2019sophie} have been employed for this task. In \cite{zhang2019sr}, LSTMs were used for single pedestrian modeling and an attention-based state refinement module designed to capture the spatial relations among different pedestrians. Graph-based attention has been proposed for spatial relation modeling \cite{yu2020spatio}, where the graph is built based on spatial distance among pedestrians. The difference between this task and ours is that the former aims to predict the future based on past observation for all pedestrians, while we focus more on models that can continually predict about how target participants will react to behavior of other observed participants. This is particularly important in activities that have very strongly coupled interactions.
Nonetheless, existing methods can be applied to our task with minor modification, as described later.

\section{Method}
\label{sec:method}

\subsection{Problem Definition}\label{sec_define}
Participants behavior inference and prediction are to estimate the behavior of a number of target participants in a group, based on information of other observed participants in that group. Supposed there are $N$ participants in the group and they are divided into two sets, with $N_{\text{obs}}$ observed participants and $N_{\text{tgt}}$ target participants, where $N$$=$$N_{\text{obs}}$$+$$N_{\text{tgt}}$. Given a trimmed video clip with $T$ frames, let $\bm{x}$$=$$\{x_{i,t}\}_{i=1:N_{\text{obs}},t=1:T}$ denote the behavior of observed participants, where the behavior comprise positions and action labels. Correspondingly, $\bm{y}$$=$$\{y_{i,t}\}_{i=1:N_{\text{tgt}},t=1:T}$ denote the behavior of target participants.

The task is to infer and predict $\{y_{i,t}\}_{i=1:N_{\text{tgt}}}$, starting from known initial states of the target participants, $\{y_{i,1}\}_{i=1:N_{\text{tgt}}}$. The estimation proceeds sequentially in time, where the observable input at time $t$ consists of $\{x_{i,\tau}\}_{i=1:N_{\text{obs}},\tau=1:t+K}$, where $K$ is the number of frames \emph{into the future} beyond $t$. Here, $K$ can be interpreted as the level of (perfect) human foresight of the target participants in predicting how other participants may behave.
As an ML problem, $K$$=$$0$ corresponds to participants behavior prediction, while it becomes inference for $K$$\geq$$1$. The inference can be performed in an online manner if $K$$=$$1$, otherwise it has to be offline or with a delay.

\subsection{Typical Transformer}\label{sec_typical}
A typical Transformer consists of multiple layers of encoder and decoder. Both encoder and decoder involve three modules: attention function, feed forward network (FFN), and normalization, where attention function is 
\begin{equation}
    \label{equ_self_dot_att}
    X^{\text{att}}=f_o\left[\frac{\bm{S}\left(f_q\left(X_q\right)f_k\left(X_k\right)^T\right)}{\sqrt{d}}f_v\left(X_v\right)\right]+X_q.
\end{equation}
In \eqref{equ_self_dot_att}, $X_q$, $X_k$, and $X_v$ denote the input feature map of query, key, and value correspondingly, and $X^{\text{att}}$ is the output attended feature map. $f\left(\cdot\right)$ is the fully-connected (FC) layer, $\bm{S}\left(\cdot\right)$ is the softmax function on each row of the input matrix, and $d$ is the dimension of $X_q$ and $X_k$. 
Noted that multi-head attention scheme in \cite{vaswani2017attention} is also employed in all attention modules of our framework, which is ignored in \eqref{equ_self_dot_att} for simplification.

A typical transformer \cite{vaswani2017attention} can fit the proposed task, since the feature of observed and target participants can be treated as two different sequences. Compared with machine translation, the observed participants sequence plays the role of source language sentence and the target participants sequence plays the role of target language sentence. However, a typical transformer has a drawback that leads to error accumulation in the task of participant behavior inference and prediction. The attention function \eqref{equ_self_dot_att} takes some other feature (key and value) into consideration when maps the input (query) to the output. From another view, the attention function can be seen as a summarization of the three entries. Different from convolutions or MLP, the three entries play different roles in the attention function. Specifically, the query is the base in the attention function while key and value are the references. In the inference stage, the query of decoder comes from the previous frame estimation, which is not accurate. With a noisy or wrong query entry, it is difficult to recover the feature and provide a relative correct estimation in the next frame. Therefore, the error will accumulate over time, which may not be as relevant in open-ended tasks, \eg text generation.

\subsection{Entry-Flipped Transformer}\label{sec_ef}
To solve the error accumulation problem, an EF-Transformer is proposed. In our EF-Transformer, encoders apply spatio-temporal attention modules to encode the information from multiple participants in the whole clip. Different from typical transformers, the decoder in EF-Transformer takes the output of the encoder as the query entry. Since this does not depend as much on predictive accuracies in previous frames, it reduces the accumulation of errors. With the Spatio-Temporal Encoder (ST-Encoder) and Entry-Flipped Decoder (EF-Decoder), the proposed EF-Transformer is designed to predict the behavior of target participants frame-by-frame more from observations rather than earlier predictions.
\begin{figure}
    \centering
    \subfigure[ST-Encoder]{\label{FigEnc}\includegraphics[width=25mm]{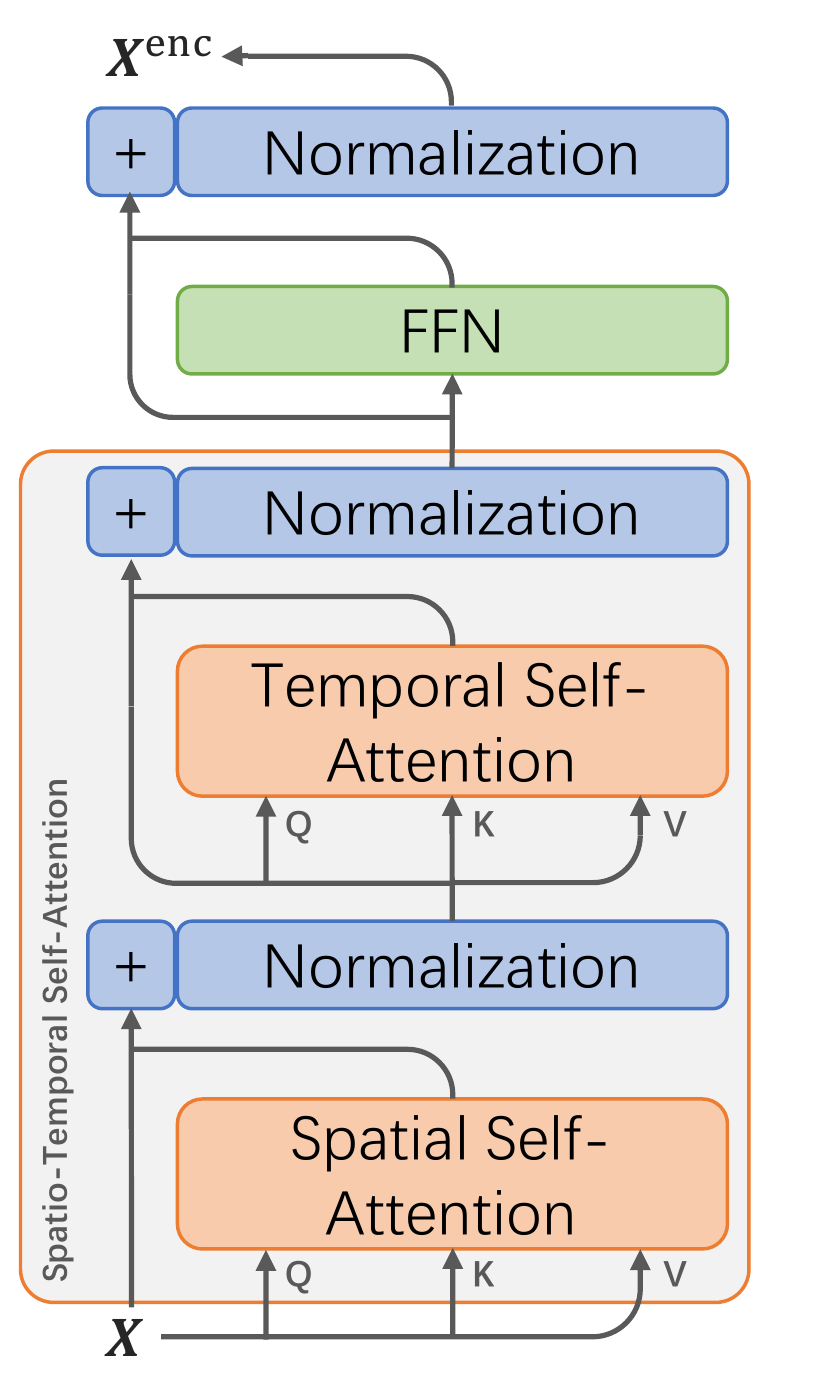}} 
    \subfigure[EF-Decoder]{\label{FigDec}\includegraphics[width=33.5mm]{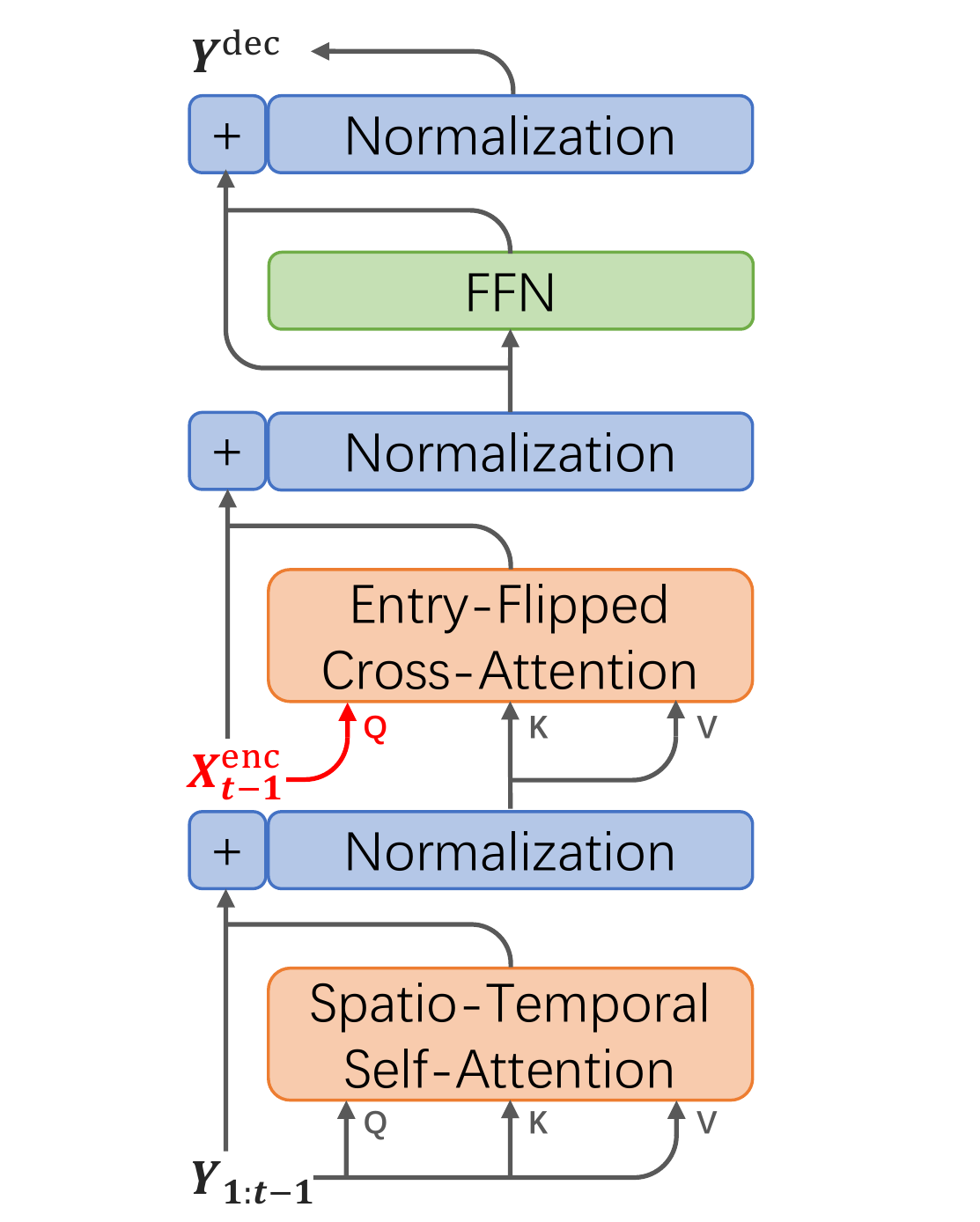}} 
    \subfigure[Prediction]{\label{FigPred}\includegraphics[width=45mm]{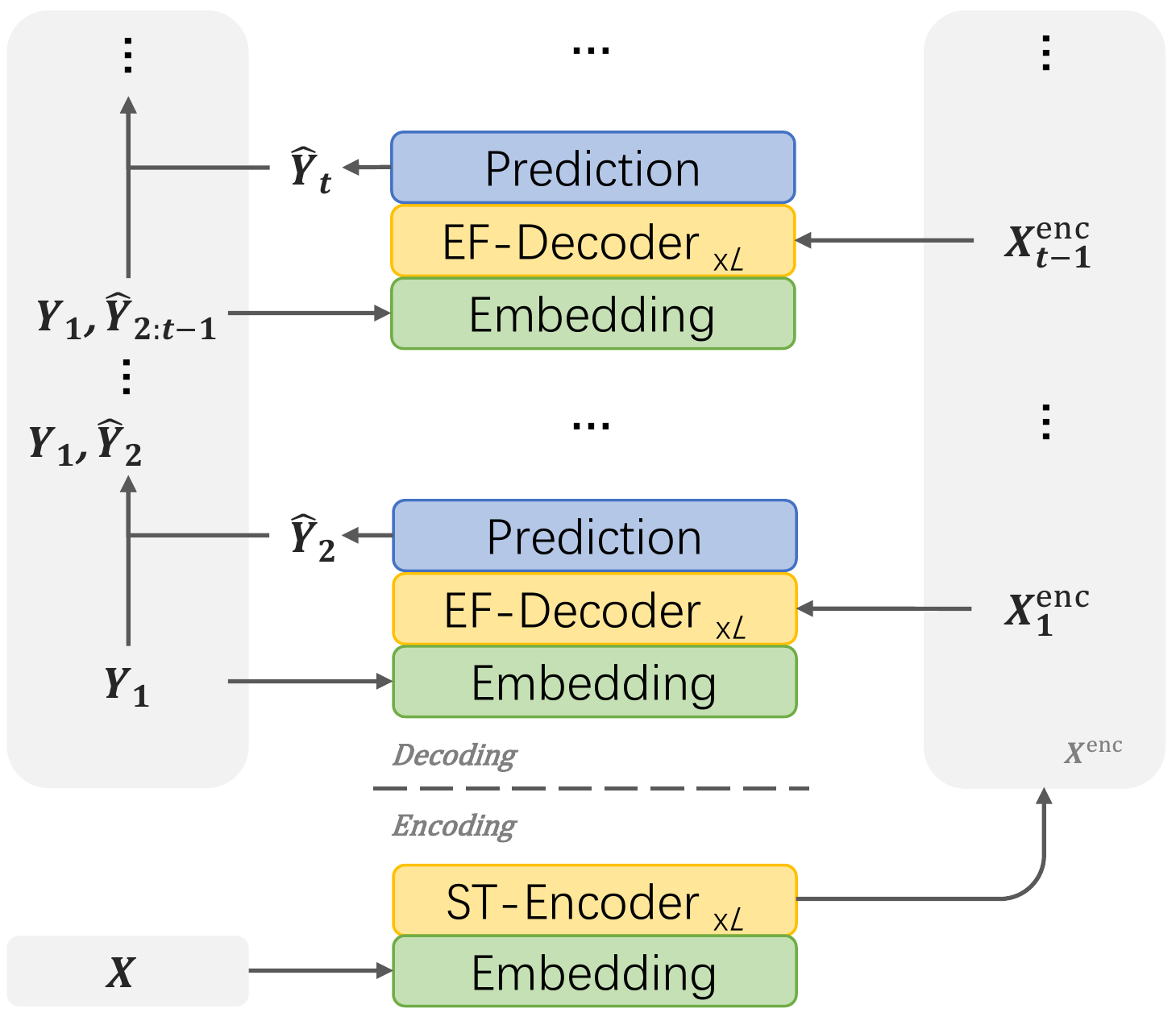}}
    \caption{The framework of encoder, decoder, and prediction process in the proposed EF-Transformer. For participants inference, $X_t^{\text{enc}}$ is sent to decoder to estimate $\hat{Y}_t$.}
    \label{fig:FigEncDec}
\end{figure}

\subsubsection{Spatio-Temporal Encoder}\label{sec_encoder}
An ST-Encoder employs two self-attention functions and an FFN to map the features of observed participants $\bm{x}$ to encoded features $\bm{x}^{\text{enc}}$, as shown in Fig.\ \ref{FigEnc}. Different from word sequences, there are both spatial and temporal domains in each video clip. 
As the attention function has a quadratic time complexity of input size \cite{vaswani2017attention}, the time complexity of an attention function over the combined spatio-temporal domain is $\mathcal{O}\left(N^2T^2\right)$. To reduce this, the attention over the two domains are handled separately. Spatial self-attention captures the relation among all participants in one frame, where every frame is sent to spatial self-attention independently. Subsequently, temporal self-attention captures the relation among all time frames for each participant to get the attended feature $\bm{x}^{\text{att}}$, so that different participants across different time frames are not directly attended. By dividing the self-attention of observed participants into two domains, the time complexity is reduced to $\mathcal{O}(NT(N+T))$.
Masked attention \cite{vaswani2017attention} is applied to avoid attending the feature beyond $K$ frames. Following \cite{vaswani2017attention}, a simple FFN is connected to the output of self-attention to obtain $\bm{x}^{\text{enc}}$ from $\bm{x}^{\text{att}}$.

\subsubsection{Entry-Flipped Decoder}\label{sec_decoder}
In the decoding stage, an EF-Decoder module is introduced. This consists of a self-attention function, a cross-attention function, and an FFN. The self-attention in EF-Decoder is also divided into spatial and temporal domains, which has the same structure as ST-Encoder. It provides the self-attended feature of target participants $\bm{y}^{\text{att}}$.
Unlike in a typical transformer, cross-attention in the proposed EF-Decoder uses as query the encoded features of observed participants, while key and value entries are self-attended features of target participants, including both those initially observed and later predicted. This is shown in Fig.\ \ref{FigDec}.
Specifically, when predicting frame $\tau$, $\{x^{\text{enc}}_{i,\tau-1}\}_{i=1:N_{\text{obs}}}$ is the query entry and $\{y^{\text{att}}_{i,t}\}_{i=1:N_{\text{tgt}},t=1:\tau-1}$ form the key and value entries. The key idea is that the \emph{query only contains observed participants in the current frame}, which becomes the base for next frame inference or prediction. Keys and values only contain target participants in past frames, forming the reference bases for next frame inference or prediction. The decoded feature $\bm{y}^{\text{dec}}$ comes from an FFN stack on the cross-attention function, which is the same as the ST-Encoder.

\noindent\textbf{Justification of Entry Flipping.}
Why is this difference between our method and a typical transformer important?
For NLP translation, the most crucial word usually is the last translated word. Hence, a typical transformer uses the last translated word in the target language as the query entry of cross-attention in the decoder. However, \emph{in scenarios where the behavior of participants are highly coupled and reactive}, such as in game sports, the most important clue for determining the behavior of a target participant in next frame would \emph{not be the past frames of the participant}, but rather the status of \emph{other observed participants in the current frame}.
For example, the ideal movement of a tennis player highly depends on the evolving positions of her teammate and opponents, whereas rapid acceleration and direction changes mean that the historical positions of this player is not that critical as a predictor. Therefore entry-flipping is more appropriate for the proposed group behavior inference and prediction tasks.

\subsubsection{Prediction Framework}\label{sec_prednet}
The whole prediction (Fig.\ \ref{FigPred}) network includes several layers: i) feature embedding layer, ii) ST-Encoder layers, iii) EF-Decoder layers, and iv) prediction layer.

\noindent\textbf{Feature Embedding.}
Two FC layers are separately applied on the two types of input, \ie 2D coordinates and action labels of participants, to map to higher dimensional features.
We first expand the 2D coordinates $\left(u_{i,t},v_{i,t}\right)$, to a normalized 7D geometric feature $x_{i,t}^{\text{g}}$ by 
\begin{equation}
    \label{equ_geo}
    x_{i,t}^{\text{g}} = \left[uv_{i,t},uv^{\Delta}_{i,t},uv^{R}_{i,t},t/T\right]^T,
\end{equation}
where
\begin{equation}
    \label{equ_7d}
    \begin{split}   
        uv_{i,t}= & \left[\frac{u_{i,t}}{w},\frac{v_{i,t}}{h}\right], \\  
        uv^{\Delta}_{i,t}= & \left[\frac{u_{i,t}-u_{i,t-1}}{w},\frac{v_{i,t}-v_{i,t-1}}{h}\right], \\  
        uv^{R}_{i,t}=&\left[\frac{u_{i,t}-u_{i,1}}{w},\frac{v_{i,t}-v_{i,1}}{h}\right] \\  
    \end{split}
\end{equation}
for a video frame of width $w$ and height $h$, for which $x_{i,t}^{\text{g}}$ contain absolute coordinates, relative coordinates, and temporal positions, all of which are normalized. $x_{i,t}^{\text{g}}$ is sent to a FC layer $f_g$ to obtain higher dimensional geometric features. Action labels are first converted to one-hot $x_{i,t}^{\text{s}}$, followed by another FC layer $f_s$. Both types of features are concatenated before positional encoding $x_{i,t}^{\text{pe}}$ \cite{vaswani2017attention} is added. Thus, the feature of a participant is 
\begin{equation}
    \label{equ_map}
    x_{i,t} =  \left[
    \begin{matrix}
        f_g(x_{i,t}^{\text{g}}), f_s(x_{i,t}^{\text{s}})
    \end{matrix}
    \right]^T+x_{i,t}^{\text{pe}}.
\end{equation}

\noindent\textbf{Encoders and Decoders.} $L$ layers of ST-Encoder and EF-Decoder are stacked. The encoded feature of observed participants from output of last layer ST-Encoder is used as the query entry of all layers of EF-Decoder. The last EF-Decoder layer output is the feature that ready for target participants inference and prediction.

\noindent\textbf{Prediction.} A prediction layer provides a mapping of $\mathbb{R}^{N_{\text{obs}}\times D}\mapsto\mathbb{R}^{N_{\text{tgt}}\times D_{\text{out}}}$, where $D$ is the feature dimension of one participant in one frame. The features of $N_{\text{obs}}$ observed participants are flattened before inference or prediction. $D_{\text{out}}$ is dimension of output, which is 2 for trajectory estimation and number of action categories for action classification. The prediction layer consists of three FC layers, where every layer is followed by a nonlinear layer (LeakyReLU in our experiment) except the last. More implementation details can be found in the supplementary.

\noindent\textbf{Loss function.}
This is a simple L2 loss applied to both trajectory and action estimation:
\begin{equation}
    \label{equ_loss}
    L = \sum_{i=1}^{N_{\text{tgt}}}\sum_{t=2}^{T}\left\|x^{\text{g*}}_{i,t}-\hat{x}^{\text{g*}}_{i,t}\right\|_2+\lambda\left\|x^{\text{s}}_{i,t}-\hat{x}^{\text{s}}_{i,t}\right\|_2,
\end{equation}
where $x^{\text{g*}}_{i,t}$ excludes the temporal coordinates ${t/T}$ in $x^{\text{g}}_{i,t}$ of \eqref{equ_geo}. In all our experiments, $\lambda$$=$$0.1$.

\section{Experiments}

\subsection{Datasets and Metrics}
We selected three datasets with closely coupled behavior in experiments. 

\noindent\textbf{Tennis Dataset} A new tennis doubles dataset was collected to evaluate our method. There are 12 videos of whole double games with resolution of $1280\times720$. 4905 10-frame clips were collected in total, which are downsampled to 2.5 fps and stabilized to remove camera motion. Individual-level bounding boxes and action labels were annotated, with the bottom-center point of each box representing the spatial location of the player. Coarse spatial positions of the ball were also estimated. As it is difficult to determine due to extreme motion blur when the ball was traveling fast, the ball position was only coarsely estimated by spatio-temporal linear interpolation between the locations of two players consecutively hitting the ball. Detailed information of the tennis dataset can be found in the supplemental material. In our experiments, the top-left player was selected as the target participant during testing, while the other three players and the ball were treated as observed entities.

\noindent\textbf{Dance Dataset} The dance dataset \cite{aizeboje2016ceilidh} contains 16 videos from overhead view of Ceilidh dances by two choreographers, where every dance was performed by 10 dancers. Two videos for each choreographer were selected for testing and others for training. The raw video is 5 fps and resolution is $640$$\times$$480$. Here 3754 10-frame clips were collected. The action labels are defined as `stand', `walk left', `walk right', `walk up',`walk down', and `twirling'. 
No explicit information about the choreographer was provided during training. 

\noindent\textbf{NBA Dataset} NBA dataset \cite{yue2014learning} contains players and ball tracking data from basketball games. During pre-processing, frame rate was down-sampled to 6.25 fps and a subset of over 4000 10-frame clips was built. As actions are not provided in this dataset, we simply assigned `defensive' and `offensive' to players as action labels. During training, one defensive player is randomly selected as the target participant, while the first defensive player in the list is selected in testing. The `resolution' (or court size) in this dataset is $100$$\times$$50$. 

\noindent\textbf{Pedestrian Datasets} ETH \cite{Lerner2007crowd} and UCY \cite{pellegrini2009you} datasets are conventionally used in pedestrian trajectory prediction. 
Target participants were randomly selected in training, and the one with longest trajectory was picked in testing. Four nearest neighbors of the target pedestrian among all frames were selected as observed participants.
We follow the leave-one-out evaluation in \cite{gupta2018social}. 

\noindent\textbf{Metrics} To evaluate the accuracy of trajectory inference and prediction, two metrics were computed following \cite{zhang2019sr}: Mean Average Displacement (MAD) is the mean distance between estimation and ground truth over all frames. Final Average Displacement (FAD) is the distance between estimation and ground truth of the last frame. Besides, metrics of short, middle, and long trajectory lengths were computed separately, where the length threshold was statistically determined over all samples to even out the number of samples across each category. For action inference and prediction, Macro F1-scores are reported.

\subsection{Baseline and Other SOTA Methods} \label{sec_baseline}
We compare with several methods in our experiments:

\noindent\textbf{CNN-based Method.} This framework is based on spatial and temporal convolutional layers. The encoder consists of 2 convolutional layers while the decoder consists of 3 convolutional layers. A 5-frame sliding window is applied for input. 

\noindent\textbf{RNN-based Method.} This framework has encoders and decoders based on two GRU layers. At each frame, the output of the encoder is concatenated with the historical data of target participants before sending to the decoder.

\noindent\textbf{Typical Transformer.} The typical transformer \cite{vaswani2017attention} here uses the ST-encoder and a typical decoder structure, with an additional future mask added to the attention function of encoding stage.

\noindent\textbf{Pedestrian Trajectory Prediction Methods.} \cite{yu2020spatio,zhang2019sr} are also compared. Modifications are made to apply them to our tasks: i) ground truth of observed pedestrians are provided for all frames in the testing stage, ii) if $K$$>$$0$, a $K$-frame time shift over target participants is adopted to ensure the network has correct information of $K$-frame future of observed participants.

\subsection{Ablation Study}
\begin{figure}[tb]
    \centering
    \begin{minipage}[t]{0.24\textwidth}
        \centering
        \includegraphics[width=3cm]{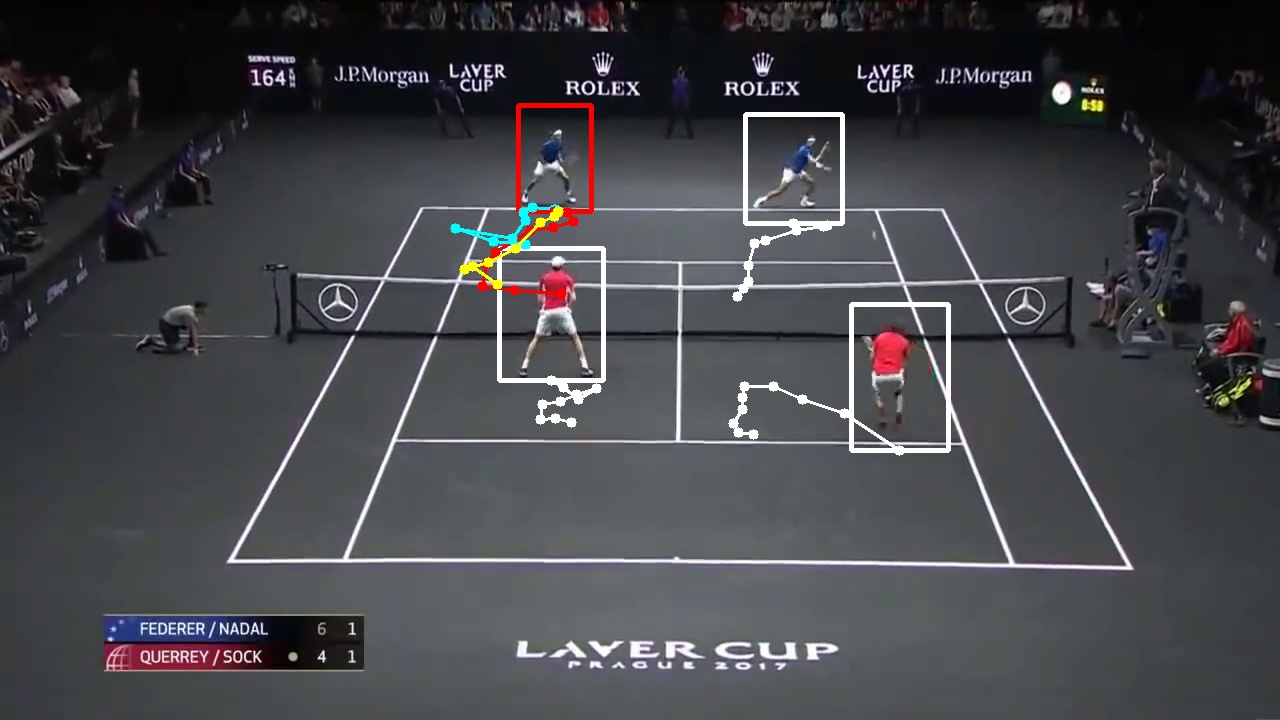}
        \centerline{}
    \end{minipage}
    \begin{minipage}[t]{0.24\textwidth}
        \centering
        \includegraphics[width=3cm]{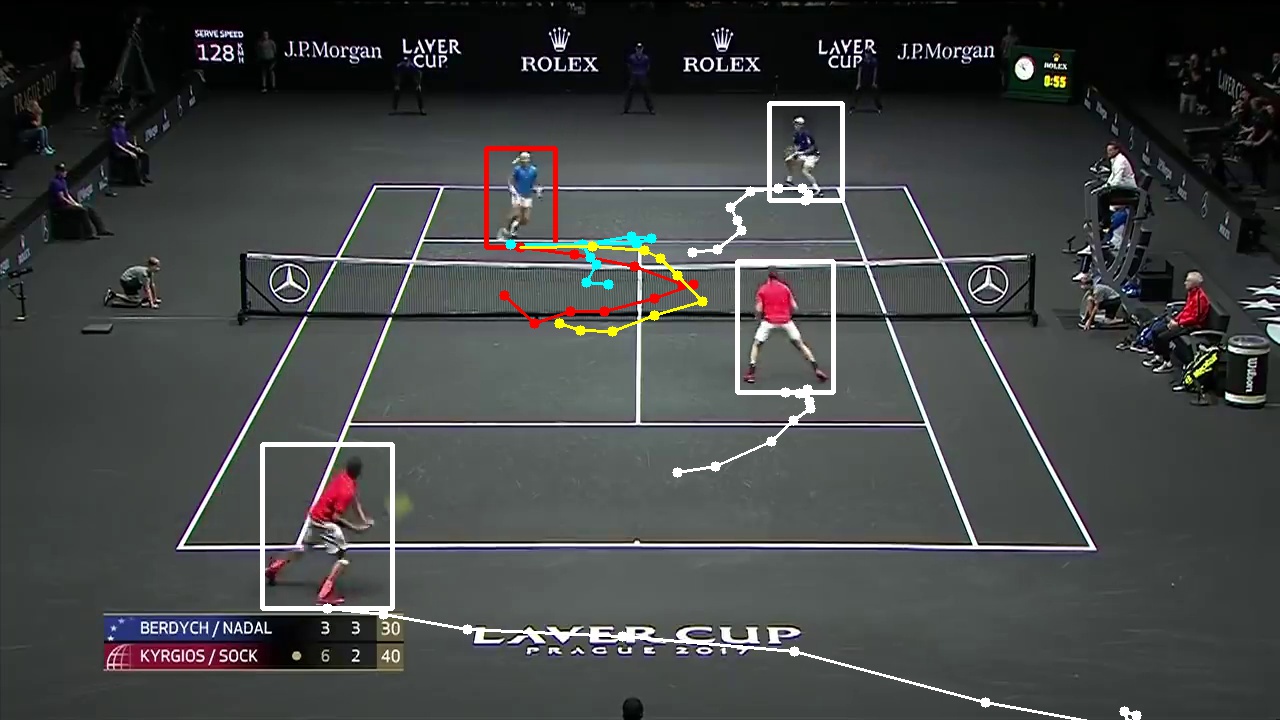}
        \centerline{} 
    \end{minipage}
    \begin{minipage}[t]{0.24\textwidth}
        \centering
        \includegraphics[width=3cm]{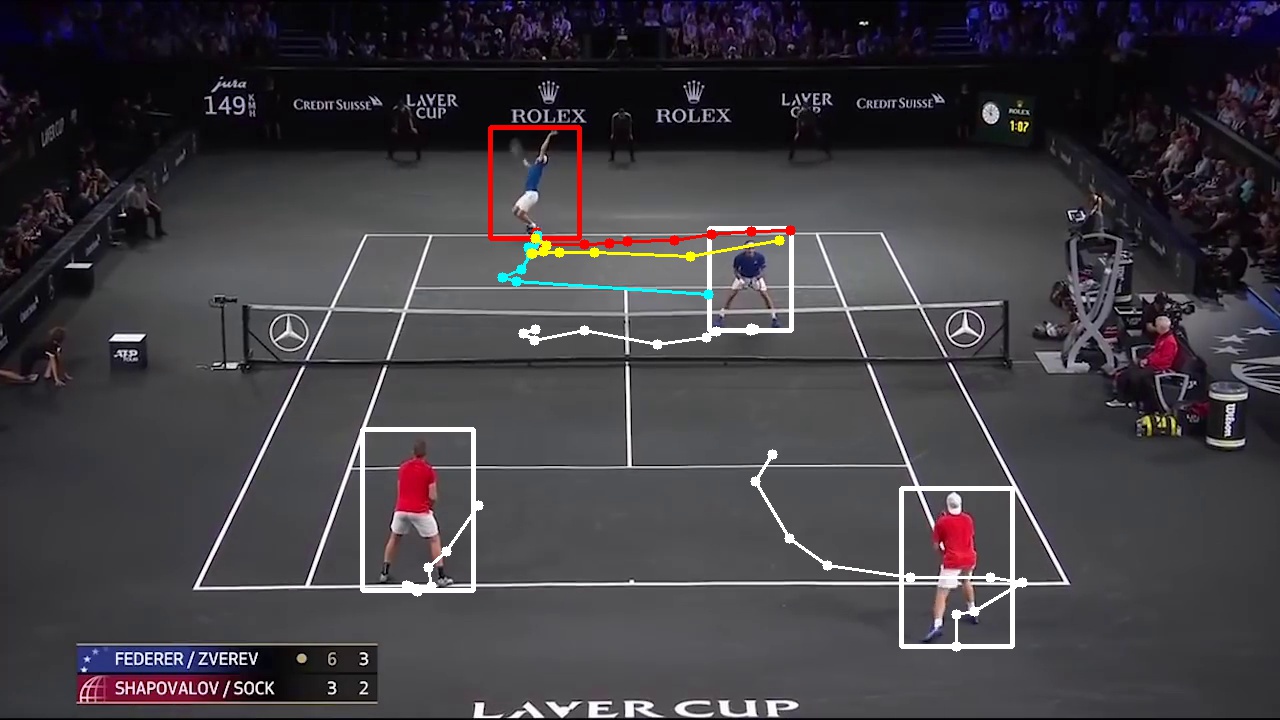}
        \centerline{}
    \end{minipage}
    \begin{minipage}[t]{0.24\textwidth}
        \centering
        \includegraphics[width=3cm]{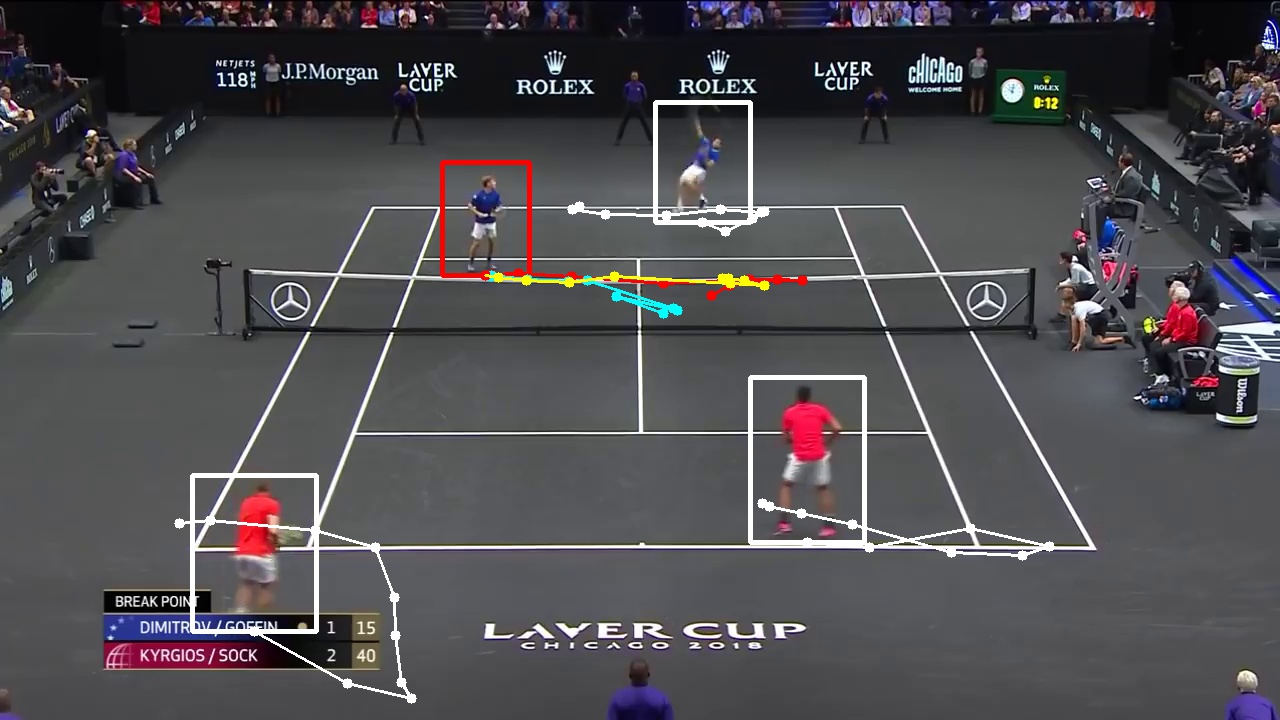}
        \centerline{} \\
    \end{minipage}
    \caption{Visualization of trajectory prediction results of EF-Transformer and typical transformer on tennis dataset. White rectangles and trajectories are the observed participants. Red rectangles are target participants with red trajectories for ground truth. Cyan trajectories are predicted by typical transformer and yellow ones are predicted by our method. Please zoom in to see details.}   
    \label{fig_tennisvis}
\end{figure}
\begin{table}[ht]
    \renewcommand\arraystretch{1.23}
    \setlength\tabcolsep{3pt}
    \caption{Comparisons of different ST-Encoders and prediction types on Tennis dataset.} 
    \label{tab_ablation}
    \begin{center}
        \scriptsize
        \begin{tabular*}{0.88\textwidth}{@{\extracolsep{\fill}}c | c | c  c  c  c | c  c  c  c  }
            \hline
            \multirow{2}*{\textbf{Encoder}} & 	\multirow{2}*{\textbf{Pred}}&  \multicolumn{4}{c|}{\textbf{MAD}} & \multicolumn{4}{c}{\textbf{FAD}} \\
            \cline{3-10}
            ~ & ~ 	& \textbf{Short} & \textbf{Mid} & \textbf{Long} & \textbf{Avg} & \textbf{Short} & \textbf{Mid} & \textbf{Long} & \textbf{Avg} \\
            \hline\hline            
            \textbf{S$+$T}		&  		& \textbf{18.70} & 31.27 & 44.51 & 28.93 & \textbf{32.89} & 51.56 & 69.15 & 47.74 \\
            \textbf{T$\rightarrow$S}	& $uv^{R}$		& 19.49 & 31.01 & 45.71 & 29.29 & 35.31 & 50.96 & 69.81 & 48.43 \\
            \textbf{S$\times$T}		&  		& 19.72 & 32.05 & 43.73 & 29.53 & 36.28 & 54.09 & 67.90 & 49.95 \\
            
            \hline
            & $uv$ 						& 40.52 & 50.42 & 62.73 & 48.89 & 36.11 & 49.05 & 64.33 & 46.91 \\
            \textbf{S$\rightarrow$T} 	& $\Sigma uv^{\Delta}$	& 20.72 & 32.91 & 49.05 & 31.18 & 40.12 & 57.81 & 78.98 & 54.93 \\
            & $uv^{R}$		& 19.40 & \textbf{30.04} & \textbf{43.04} & \textbf{28.35} & 35.38 & \textbf{48.62} & \textbf{64.23} & \textbf{46.43}      \\ 
            \hline
        \end{tabular*}
    \end{center}
\end{table}
In this section, we compare several ST-Encoder structures. \textbf{S$+$T} represents the parallel structure, where spatial and temporal self-attentions are operated on separately, with the outputs added together. \textbf{S$\rightarrow$T} and \textbf{T$\rightarrow$S} are sequential structures with different order of spatial and temporal domain. \textbf{S$\times$T} represents jointly computing attention functions over spatial and temporal domain.
In addition, we evaluated the accuracy of different position estimators from among the 3 predicted components in \eqref{equ_7d}, which have overlapping redundancy. Here, the frame-wise relative component $uv^{\Delta}$ is cumulatively summed to get position estimates $\Sigma{uv^{\Delta}}$, relative to target positions in the last fully-observed frame. Results are shown in Table \ref{tab_ablation}.

Of the three components, $uv^{R}$ appeared to be better predicted than the other two. Prediction of absolute coordinates $uv$ is more difficult than only predicting the difference. However, predicting the difference of neighboring frames $uv^{\Delta}$ suffers from error accumulation. The output of frame $t$ have to compensate for the error in predicting frame $t$$-$$1$, which can lead to unstable oscillations. Compared with Parallel ST-Encoder, Sequential ST-Encoder achieved better performance except on short trajectories. This is because the query of Sequential ST-Encoder is capable of attending to all other participants in all frames, while query of parallel encoders can only attend the same participants in different frames, or other participants in the same frame. Based on the results above, only predictions of $uv^{R}$ are reported in the following experiments.

\subsection{Trajectory Inference and Prediction}
\begin{table}[htb]
    \renewcommand\arraystretch{1.23}
    \setlength\tabcolsep{3pt}
    \caption{Comparisons of trajectory inference and prediction with baselines and SOTA methods on tennis dataset and NBA dataset.} 
    \label{tab_tennis}
    \begin{center}
        \scriptsize
        \begin{tabular*}{0.9\textwidth}{@{\extracolsep{\fill}} c | c | c | c  c  c  c | c  c  c  c  }
            \hline
            \multicolumn{2}{c|}{} & 	\multirow{2}*{\textbf{Methods}}&  \multicolumn{4}{c|}{\textbf{MAD}} & \multicolumn{4}{c}{\textbf{FAD}} \\
            \cline{4-11}
            \multicolumn{2}{c|}{}  & ~ 	& \textbf{Short} & \textbf{Mid} & \textbf{Long} & \textbf{Avg} & \textbf{Short} & \textbf{Mid} & \textbf{Long} & \textbf{Avg} \\
            \hline\hline
            \multirow{6}*{\rotatebox{90}{\textbf{Inference}}} & \multirow{12}*{\rotatebox{90}{\textbf{Tennis}}}
            & CNN-based 					& 22.61 & 41.63 & 64.43 & 38.54 & 42.97 & 73.27 &102.78 & 67.22 \\
            ~&~& RNN-based					& 22.62 & 41.27 & 72.88 & 39.78 & 38.07 & 67.86 &103.47 & 63.01 \\
            ~&~& Transformer					& 21.17 & 32.91 & 46.67 & 30.95 & 37.14 & 52.06 & 68.14 & 49.34 \\
            ~&~& SR-LSTM \cite{zhang2019sr} 	& 21.22 & 34.46 & 55.60 & 33.19 & 41.49 & 58.50 & 90.08 & 57.60 \\
            ~&~& STAR \cite{yu2020spatio}		& 20.28 & 35.21 & 55.36 & 33.16 & 36.86 & 57.52 & 90.01 & 55.45 \\
            ~&~& EF-Transformer				& \textbf{19.40} & \textbf{30.04} & \textbf{43.04} & \textbf{28.35} & \textbf{35.38} & \textbf{48.62} & \textbf{64.23} & \textbf{46.43} \\ 
            \cline{1-1}\cline{3-11}
            \multirow{10}*{\rotatebox{90}{\textbf{Prediction}}} &~
            &   CNN-based 					& 22.58 & 41.81 & 71.57 & 39.80 & 38.84 & 70.35 &105.26 & 64.76 \\
            ~&~& RNN-based					& 23.84 & 41.99 & 78.97 & 41.57 & 41.34 & 68.29 &110.63 & 65.58 \\
            ~&~& Transformer					& 20.14 & 33.09 & 50.70 & 31.33 & 35.85 & 52.55 & 71.57 & 49.67 \\
            ~&~& SR-LSTM \cite{zhang2019sr} 	& 20.43 & 43.86 & 85.88 & 42.37 & 39.11 & 75.36 &117.43 & 69.25 \\
            ~&~& STAR \cite{yu2020spatio}		& 23.83 & 43.80 & 83.65 & 43.20 & 37.83 & 70.61 &117.19 & 66.50 \\
            ~&~& EF-Transformer				& \textbf{19.24} & \textbf{30.71} & \textbf{41.98} & \textbf{28.44} & \textbf{34.97} & \textbf{50.36} & \textbf{62.60} & \textbf{46.83} \\ 
            \cline{2-11}
            ~ &\multirow{4}*{\rotatebox{90}{\textbf{NBA}}}&Transformer				& 1.78 & 4.25 & 10.13 & 3.99 & 2.91 & 7.33 & 18.14 & 6.93  \\ 
            ~&~& SR-LSTM \cite{zhang2019sr}			& 2.84 & 4.78 & 10.53 & 4.77 & 6.00 & 8.90 & 18.63 & 9.08\\ 
            ~&~& STAR \cite{yu2020spatio}	    	& 4.51 & 5.96 & \textbf{10.04} & 5.92 & 5.81 & 8.81 & 18.07 & 8.86 \\ 
            ~&~& EF-Transformer				& \textbf{1.65} & \textbf{4.18} & 10.05 & \textbf{3.89} & \textbf{2.69} & \textbf{7.23} & \textbf{18.00} & \textbf{6.75} \\ 
            \hline
        \end{tabular*} 
    \end{center}
\vspace{-15pt}
\end{table}
Here we focus solely on trajectory estimation, so ground truth action labels were provided for target participants. Table \ref{tab_tennis} presents the results of behavior prediction and inference on the tennis and NBA datasets. For the tennis dataset, it can be observed that our EF-Transformer achieved the best performance among compared methods, in particular significantly outperforming other methods for long trajectories. Longer trajectories provide greater risk of larger estimation errors, and our entry-flipping mechanism is effective for limiting error accumulation. Performance of SR-LSTM \cite{zhang2019sr} is affected by the limited initial ground truth sequence of target participants to adequately bootstrap the LSTM cell states. Furthermore, estimated coordinates of target participants are sent to the state refinement module, so the hidden state of observed participants may become affected by past estimation errors. Similarly, STAR \cite{yu2020spatio} models the spatial relations of all participants together, where the features of observed participants will also become conflated with inferred features of target participants. Comparing inference and prediction, prediction is harder for all methods as no future information of observed participants is provided. This is especially in the tennis dataset, where the behavior of target participants involve quick reactions to observed participants, often with anticipatory foresight. Some visualizations are shown in Fig.\ \ref{fig_tennisvis}, illustrating that our method can predict better trajectories than a typical transformer. 

In the NBA dataset, EF-Transformer also outperformed other methods except for the MAD of long trajectories, where STAR \cite{yu2020spatio} surpassed ours only by a tiny 0.01. It can be observed from Table \ref{tab_tennis} that the performance differences among compared methods are less than for the tennis dataset.
We believe the main reason is that in the most of the cases, a defensive player only needs to follow the corresponding offensive player, which is a simpler reaction than the tennis scenario and usually results in a small displacement during prediction for all methods.
\begin{table}[htb]
    \renewcommand\arraystretch{1.23}
    \setlength\tabcolsep{3pt}
    \caption{Comparisons of trajectory prediction with 1 and 2 target participants with baselines and SOTA methods on dance dataset.} 
    \label{tab_dance}
    \begin{center}
        \scriptsize
        \begin{tabular*}{0.88\textwidth}{@{\extracolsep{\fill}}c | c | c  c  c  c | c  c  c  c  }
            \hline
            \multirow{2}*{ } & 	\multirow{2}*{\textbf{Methods}}&  \multicolumn{4}{c|}{\textbf{MAD}} & \multicolumn{4}{c}{\textbf{FAD}} \\
            \cline{3-10}
            ~ & ~ 	& \textbf{Short} & \textbf{Mid} & \textbf{Long} & \textbf{Avg} & \textbf{Short} & \textbf{Mid} & \textbf{Long} & \textbf{Avg} \\
            \hline\hline
            \multirow{6}*{\rotatebox{90}{\textbf{$N_{\text{tgt}}$$=$$1$ }}}
            & CNN-based 					&  6.91 & 12.19 & 14.58 & 11.13 &  8.64 & 14.49 & 16.86 & 13.22 \\
            ~& RNN-based					&  8.60 & 15.09 & 20.52 & 14.61 & 10.71 & 17.08 & 20.69 & 16.03 \\
            ~& Transformer					&  7.29 & 12.75 & 17.33 & 12.35 &  9.63 & 14.83 & 19.43 & 14.53 \\
            ~& SR-LSTM \cite{zhang2019sr} 	&  9.50 & 15.67 & 22.48 & 15.76 & 11.56 & 18.16 & 21.82 & 17.05 \\
            ~& STAR \cite{yu2020spatio}		&  9.25 & 15.34 & 22.34 & 15.52 & 11.76 & 18.93 & 23.70 & 17.99 \\
            ~& EF-Transformer				&  \textbf{6.28} & \textbf{9.99} & \textbf{12.11} & \textbf{9.39}  & \textbf{7.42} & \textbf{10.83} & \textbf{12.56}  & \textbf{10.20} \\ 
            \hline
            \multirow{6}*{\rotatebox{90}{\textbf{$N_{\text{tgt}}$$=$$2$}}}
            & CNN-based 					&  7.24 & 12.55 & 14.99 & 11.49 &  8.78 & 14.86 & 16.97 & 13.42 \\
            ~& RNN-based					&  9.20 & 15.77 & 20.93 & 15.17 & 11.56 & 17.72 & 21.47 & 16.79 \\
            ~& Transformer					&  7.02 & 12.26 & 17.49 & 12.15 &  9.39 & 15.50 & 20.06 & 14.86 \\
            ~& SR-LSTM \cite{zhang2019sr} 	&  9.19 & 13.92 & 18.21 & 13.68 & 10.69 & 15.09 & 18.11 & 14.54 \\
            ~& STAR \cite{yu2020spatio}		&  8.26 & 14.78 & 22.77 & 15.14 & 10.39 & 17.07 & 23.34 & 16.80 \\
            ~& EF-Transformer				&  \textbf{6.80} & \textbf{10.19} & \textbf{12.23} &  \textbf{9.67} &  \textbf{8.22} & \textbf{11.52} & \textbf{13.60} & \textbf{11.05} \\ 
            \hline
        \end{tabular*} 
    \end{center}
    \vspace{-15pt}
\end{table}

For the dance dataset, we evaluated the methods on a prediction task with different numbers of target participants. Results are listed in Table \ref{tab_dance}. Our method outperformed all compared methods. It can also be observed that the results of $N_{\text{tgt}}$$=$$2$ are comparable to $N_{\text{tgt}}$$=$$1$. Although fewer observed participants may make the prediction more difficult, it is possible that having more target participants during training provide better guidance to the network, so that the patterns of the dances are better learned. More results of inference task can be found in the supplemental material.

To evaluate the performance in pedestrian datasets, we follow the setting in \cite{gupta2018social} to provide 8-frame ground truth for the target participant, as the behavior of a pedestrian highly relies on self intention, which underlies one's historical trajectory.
The results are shown in Table \ref{tab_pedestrian}. Our method achieved the best performance among compared methods. As before, existing methods \cite{zhang2019sr,yu2020spatio} are not appropriately designed for scenarios with different sets of observed and target participants, conflating accurate observations with inaccurate past estimates. Behavior prediction with 1-frame observation for the target is also evaluated. Results and visualizations can be found in the supplementary.

\begin{table}[htb]
    \renewcommand\arraystretch{1.2}
    \setlength\tabcolsep{3pt}
    \caption{Comparisons of trajectory prediction with baselines and SOTA methods on pedestrian dataset.} 
    \label{tab_pedestrian} 
    \begin{center}
        \scriptsize
        \begin{tabular*}{0.88\textwidth}{@{\extracolsep{\fill}}c | c  c  c  c  c  c  }
            \hline
            \multirow{2}*{\textbf{Methods}} & \multicolumn{6}{c}{\textbf{Performance MAD/FAD}} \\
            \cline{2-7}
            ~ & ETH & HOTEL & ZARA & ZARA2 & UNIV & AVG \\ 
            \hline\hline             
            SR-LSTM \cite{zhang2019sr}  & 1.09/1.76&0.69/1.31&0.79/1.70&0.88/1.85&1.23/2.32&0.94/1.79 \\
            STAR \cite{yu2020spatio}	& 1.09/2.85&0.69/1.41&0.91/2.08&1.27/2.92&1.00/2.18&0.99/2.23 \\
            Transformer				    & 0.73/1.40&0.52/0.93&0.63/1.24&0.68/1.46&1.00/1.96&0.71/1.40 \\
            EF-Transformer				& \textbf{0.70}/\textbf{1.33}&\textbf{0.49}/\textbf{0.84}&\textbf{0.53}/\textbf{1.07}&\textbf{0.54}/\textbf{1.10}&\textbf{0.89}/\textbf{1.75}&\textbf{0.63}/\textbf{1.22}\\
            
            \hline
        \end{tabular*} 
    \end{center}
\vspace{-15pt}
\end{table}

\begin{table}[htb]
    \renewcommand\arraystretch{1.4}
    \setlength\tabcolsep{3pt}
    \caption{Comparisons of multi-task prediction with baselines and SOTA methods on dance dataset.`Traj' represents the task of trajectory prediction, during which ground truth action labels are provided. `Multi' represents the task of multi-task prediction, where both trajectories and action labels have to be predicted.} 
    \label{tab_multitask}
    \begin{center}
        \scriptsize
        \begin{tabular*}{0.88\textwidth}{@{\extracolsep{\fill}}c | c | c  c  c  c | c  c  c  c  }
            \hline
            \multirow{2}*{\rotatebox{90}{\textbf{ }}} & 	\multirow{2}*{\textbf{Methods}}&  \multicolumn{4}{c|}{\textbf{MAD}} & \multicolumn{4}{c}{\textbf{FAD}} \\
            \cline{3-10}
            ~ & ~ 	& \textbf{Short} & \textbf{Mid} & \textbf{Long} & \textbf{Avg} & \textbf{Short} & \textbf{Mid} & \textbf{Long} & \textbf{Avg} \\
            \hline\hline
            \multirow{2}*{\rotatebox{90}{\textbf{Traj}}}
            & Transformer				&  7.29 & 12.75 & 17.33 & 12.35 &  9.63 & 14.83 & 19.43 & 14.53 \\
            ~& EF-Transformer			&  \textbf{6.28} & \textbf{9.99} & 12.11 & \textbf{9.39}  & \textbf{7.42} & \textbf{10.83} & 12.56  & \textbf{10.20} \\ 
            \hline
            \multirow{2}*{\rotatebox{90}{\textbf{Multi}}}
            & Transformer				&  7.91 & 14.73 & 19.24 & 13.82 & 10.77 & 17.86 & 21.94 & 16.72 \\
            ~& EF-Transformer			&  6.98 & 10.31 & \textbf{11.80} & 9.63 &  8.28 & 11.65 & \textbf{12.51} & 10.75 \\
            \hline
        \end{tabular*} 
    \end{center}
    \vspace{-15pt}
\end{table}

\subsection{Multi-Task Inference and Prediction}
In multi-task inference and prediction, trajectories and action labels are estimated simultaneously. Different from previous experiments, estimated action labels are sent to feature embedding for next-frame inference or prediction. We only compare to a typical transformer on dance dataset here. As action labels are very tightly coupled between observed and target players in tennis, it turned out that both methods resulted in 100\% action classification and only minor differences to trajectory prediction in Table~\ref{tab_tennis}, hence results are placed in the supplemental material.

Trajectory prediction results are shown in Table \ref{tab_multitask}. Without ground truth action labels for target participants, our method achieved comparable trajectory prediction performance to results with ground truth input. In contrast, the typical transformer had worse performance when action labels for target participants had to be estimated. Action prediction confusion matrices are provided in the supplementary.  The macro F1-score of our method and typical transformer are 0.99 and 0.90 correspondingly. As our method is capable of limiting accumulated errors, trajectory and action predictions occur in a virtuous cycle, where error robustness in the previous step improves action classification, which in turn improves trajectory prediction. This contrasts with a typical transformer, where error drift leads to poorer action classification and larger errors in trajectory prediction.

\subsection{Robustness Analysis}
Robustness reflects the ability to limit error accumulation, as well as to recover from large errors (\eg due to sensing failure). To evaluate robustness, the 6D prediction of one middle frame is replaced by a large noise spike of [1,1,-1,-1,-1,-1]. FAD was then computed to compare how well the methods recovered from the spike. This experiment was performed with the inference task on the tennis dataset, where the spike was added to different frames.

\begin{table}[htb]
    \renewcommand\arraystretch{1.23}
    \setlength\tabcolsep{3pt}
    \caption{Comparisons of FAD on tennis dataset with noise involved in different frames.} 
    \label{tab_robust}
    \begin{center}
        \scriptsize
        \begin{tabular*}{0.88\textwidth}{@{\extracolsep{\fill}} c | c  c  c  c | c  c  c  c  }
            \hline
            \multirow{2}*{\textbf{\makecell*[c]{Noise \\ Position}}}&  \multicolumn{4}{c|}{\textbf{Transformer FAD}} & \multicolumn{4}{c}{\textbf{EF-Transformer FAD}} \\
            \cline{2-9}
            ~ 	& \textbf{Short} & \textbf{Mid} & \textbf{Long} & \textbf{Avg} & \textbf{Short} & \textbf{Mid} & \textbf{Long} & \textbf{Avg} \\
            \hline\hline
            No Noise	& 37.14 & 52.06 & 68.14 & 49.34 & 35.38 & 48.62 & 64.23 & 46.43 \\
            Noise@t=3	& 75.99 &103.24 &141.06 & 99.67 & 37.23 & 56.37 & 84.65 & 54.15 \\ 
            Noise@t=6	& 80.03 &105.35 &145.39 &105.85 & 55.19 & 64.90 & 90.71 & 65.68 \\
            Noise@t=9	&131.76 &161.07 &205.26 &157.81 &115.93 &123.30 &145.31 &124.29 \\
            \hline
        \end{tabular*} 
    \end{center}
    \vspace{-15pt}
\end{table}

Table \ref{tab_robust} shows that both methods can recover from the spike to some extent, noting that better recovery was made by the final frame for earlier spikes. Nonetheless, our method performed significantly better than the typical transformer. Even with a frame 9 spike (second-last frame), our method's FAD increased only about 78 pixels, compared to 108 pixels for the typical transformer.

\subsection{Limitations}
Our method assumes that a group has a fixed number of participants, all with strongly coupled behavior. Thus in \eg a pedestrian scenario with varying numbers of individuals, not all of whom have correlated behavior, we need to select a fixed number of the most likely related individuals as observations for each target pedestrian (\eg with k-nearest-neighbor filtering). Furthermore, although pedestrian trajectories are smoother than in tennis and dance, it turned out that prediction is also more difficult for our method. This is likely due to less behavioral coupling among pedestrians.
When observations are not as informative,
our method was predominantly trying to do some form of dead reckoning like other methods, which is difficult to be accurate especially for longer intervals.

\section{Conclusion}
In this paper, we proposed the EF-Transformer for behavior inference and prediction of target participants based on other observed participants. In our decoder, the order of query, key, and value entries of the cross-attention are flipped to effectively reduce error accumulation. EF-Transformer is evaluated in several experiments, where it outperformed all compared methods on the tennis, dance datasets and pedestrian datasets. Moreover, we demonstrate superior robustness to noise spikes. The framework of EF-Transformer can be used for application to learning realistic agent-based behavior in the future.

\paragraph{\textbf{Acknowledgements}}
This study is supported under the RIE2020 Industry Alignment Fund – Industry Collaboration Projects (IAF-ICP) Funding Initiative, as well as cash and in-kind contribution from Singapore Telecommunications Limited (Singtel), through Singtel Cognitive and Artificial Intelligence Lab for Enterprises (SCALE@NTU).
%
%
\bibliographystyle{splncs04}
\bibliography{egbib}

\newpage
\section*{Supplementary}
We first introduce the implementation details of our framework in Section \ref{sec_implement} and some detailed information of datasets in Section \ref{sec:dataset}, especially how the actions are defined. Then the supplementary experiment results are reported in Section \ref{sec:results}, followed by visualizations on the used datasets (Section \ref{sec:vis}).

\appendix

\section{Implementation Details} \label{sec_implement}
In our proposed framework, geometric and semantic inputs are mapped to 64D. $L$$=$$2$ layers of ST-Encoder and EF-Decoder are stacked. Inside each encoder and decoder, layer normalization is applied after every attention function and feed forward network (FFN). Multi-head attention is used with $H$$=$$8$ heads. In each attention function, the linear transformation of all query, key, and value entries are set to $\mathbb{R}^{128}\mapsto\mathbb{R}^{8}$. In the experiments on tennis and dance datasets, both trajectories and action labels are the inputs to the network, while for the pedestrian dataset, only trajectories are input.
Both prediction ($K$$=$$0$) and inference ($K$$=$$1$) tasks were evaluated. During training, the Adam optimizer is adopted with an initial learning rate of 0.001.

\section{Details of the Datasets}\label{sec:dataset}
\subsection{Actions in Tennis Dataset}\label{sec:tennis}
In the self-collected tennis doubles dataset, seven individual-level actions were labeled, which are shown in Table \ref{tab_label}. Actions 1 to 3 are performed in the serving stage of a tennis game, while actions 4 to 6 take place after the serving stage. As ball positions are coarsely estimated, the `action' label `7' is assigned to the ball to simplify the implementation. 

\begin{table}[htb]
    \renewcommand\arraystretch{1.3}
    \caption{Action labels in tennis dataset}
    \label{tab_label}
    \begin{center}
        \scriptsize
        \begin{tabular*}{0.65\textwidth}{@{\extracolsep{\fill}} c c  c }
            \hline
            Action Label & Action & Abbr. \\
            \hline
            1 & Serve ball 					 	& S \\
            2 & Waiting for teammate to serve ball 	& WTS \\
            3 &Waiting for opponent to serve ball 	    & WOS \\
            \hline
            4 &Pursue and hit ball 		  & P \\
            5 &Waiting for teammate to hit ball  & WTP \\
            6 &Waiting for opponent to hit ball 	 & WOP \\
            \hline
            0 & Background  & BG \\
            \hline
        \end{tabular*} 
    \end{center}
\end{table}

\subsection{Trajectory Length Statistics}
As the distance measurements in the datasets we used are different. NBA dataset measures distance in feet, while in tennis and dance datasets the distance is measured in pixels. Here we provide a statistic of trajectories distance in these three datasets Table \ref{tab_sta} so that the performance among different length categories and different datasets can be better understood.

\begin{table}[htb]
    \renewcommand\arraystretch{1.23}
    \setlength\tabcolsep{3pt}
    \caption{Statistics of trajectory length in pixels of tennis dataset (resolution 1920$\times$1080) and dance dataset (resolution 640$\times$480).} 
    \label{tab_sta}
    \begin{center}
        \scriptsize
        \begin{tabular*}{0.78\textwidth}{@{\extracolsep{\fill}} c | c  c  c   }
            \hline
            Tennis (1920$\times$1080) & Minimum Length & Maximum Length & Median Length\\
            \hline
            Short Trajectories	& 23.60 & 115.06 & 85.54 \\
            Middle Trajectories	& 115.30 & 230.02 & 159.93 \\
            Long Trajectories	& 230.50 & 524.00 & 279.86 \\            
            \hline
            \hline
            NBA (100$\times$50) & Minimum Length & Maximum Length & Median Length\\
            \hline
            Short Trajectories	& 0 & 5.99 & 3.16 \\
            Middle Trajectories	& 6.01 & 11.96 & 8.30 \\
            Long Trajectories	& 12.01 & 28.00 & 15.32 \\            
            \hline
            \hline
            Dance (640$\times$480) & Minimum Length & Maximum Length & Median Length\\
            \hline
            Short Trajectories	& 0 & 63.96 & 23.83 \\
            Middle Trajectories	& 64.27 & 127.95 & 94.31 \\
            Long Trajectories	& 128.13 & 422.17 & 169.68 \\            
            \hline
        \end{tabular*} 
    \end{center}
\end{table}

\section{Additional Experimental Results} \label{sec:results}
\subsection{Ablation Study of Decoders}\label{sec:ablation_decoder}
In the typical decoder, the query encompass past estimated target participants, while key \& value are the observed participants in all past frames. ``All Query'' indicates the straightforward entry-flipping, where query is based on observed participants in all past frames, while key \& values are the estimated target participants. 
``Limited Query'' is what was presented in our main paper, in which only observed participants in the current frame is present in the query.
Since two layers of decoder are applied, ``Hybrid'' means employing one typical decoder and one EF-Decoder sequentially, where ``TP$\rightarrow$EF'' means apply typical decoder first and vice versa. 
The results in Table \ref{tab_ablation_decoder} show that ``Limited Query'' achieves much better performance than ``All Query''. The reason is that most of observations in past frames are less important than current frame, such as tennis and basketball where both position and speed change rapidly. 
Both hybrid decoders cannot outperform ``Limited Query'' as every single typical decoder layer will accumulate more errors than our method. Hybrid versions can only introduce less error than typical transformer.

\begin{table}[ht]
    \renewcommand\arraystretch{1.1}
    \setlength\tabcolsep{3pt}
    \caption{Comparisons of different Decoders on Tennis dataset.} 
    \vspace{-10pt}
    \label{tab_ablation_decoder}
    \begin{center}
        \scriptsize
        \begin{tabular*}{0.88\textwidth}{@{\extracolsep{\fill}}c | c  c  c  c | c  c  c  c  }
            \hline
            \multirow{2}*{\textbf{Decoder}} &  \multicolumn{4}{c|}{\textbf{MAD}} & \multicolumn{4}{c}{\textbf{FAD}} \\
            \cline{2-9}
            ~ &  \textbf{Short} & \textbf{Mid} & \textbf{Long} & \textbf{Avg} & \textbf{Short} & \textbf{Mid} & \textbf{Long} & \textbf{Avg} \\
            \hline\hline
            Typical & 20.14 & 33.09 & 50.70 & 31.33 & 35.85 & 52.55 & 71.57 & 49.67 \\
            All Query & 20.80 & 33.05 & 46.86 & 30.92 & 38.25 & 55.50 & 70.14 & 51.71 \\
            Limited Query & \textbf{19.24} & \textbf{30.71} & \textbf{41.98} & \textbf{28.44} & \textbf{34.97} & \textbf{50.36} & \textbf{62.60} & \textbf{46.83} \\
            Hybrid(TP$\rightarrow$EF) & 19.49 & 31.01 & 45.71 & 29.29 & 35.31 & 50.96 & 69.81 & 48.43 \\            
            Hybrid(EF$\rightarrow$TP) & 19.52 & 31.53 & 43.84 & 29.24 & 35.53 & 52.62 & 70.57 & 49.43
            \\ 
            \hline
        \end{tabular*}	
        \vspace{-20pt}
    \end{center}
\end{table}

\subsection{Trajectory Inference and Prediction}\label{sec:traj}
\begin{table}[htb]    
    \renewcommand\arraystretch{1.3}
    \setlength\tabcolsep{3pt}
    \caption{Comparisons of trajectory inference with baselines and SOTA methods on dance dataset.} 
    \label{tab_dance_appendix}
    \begin{center}
        \scriptsize
        \begin{tabular*}{0.88\textwidth}{@{\extracolsep{\fill}}c | c | c  c  c  c | c  c  c  c  }
            \hline
            \multirow{2}*{\rotatebox{90}{\textbf{Task}}} & 	\multirow{2}*{\textbf{Methods}}&  \multicolumn{4}{c|}{\textbf{MAD}} & \multicolumn{4}{c}{\textbf{FAD}} \\
            \cline{3-10}
            ~ & ~ 	& \textbf{Short} & \textbf{Mid} & \textbf{Long} & \textbf{Avg} & \textbf{Short} & \textbf{Mid} & \textbf{Long} & \textbf{Avg} \\
            \hline\hline
            \multirow{6}*{\rotatebox{90}{\textbf{Inference}}}
            & CNN Based 					&  7.07 & 11.78 & 14.58 & 11.05 &  8.77 & 14.26 & 16.31 & 13.01 \\
            ~& RNN Based					&  7.89 & 13.30 & 16.91 & 12.59 &  9.76 & 15.17 & 17.78 & 14.13 \\
            ~& Transformer					&  6.97 & 11.76 & 13.51 & 10.66 &  8.89 & 14.38 & 15.26 & 12.74 \\
            ~& SR-LSTM \cite{zhang2019sr}   &  8.72 & 14.08 & 19.39 & 13.95 &  9.80 & 15.50 & 18.66 & 14.54 \\
            ~& STAR \cite{yu2020spatio}		&  7.94 & 13.83 & 20.93 & 14.11 &  9.15 & 14.81 & 21.20 & 14.94 \\
            ~& EF-Transformer				&  \textbf{6.81} &  \textbf{9.62} & \textbf{11.82} &  \textbf{9.36} &  \textbf{7.86} & \textbf{10.61} & \textbf{12.42} & \textbf{10.24} \\ 
            \hline
            \multirow{6}*{\rotatebox{90}{\textbf{Prediction}}}
            & CNN-based 					&  6.91 & 12.19 & 14.58 & 11.13 &  8.64 & 14.49 & 16.86 & 13.22 \\
            ~& RNN-based					&  8.60 & 15.09 & 20.52 & 14.61 & 10.71 & 17.08 & 20.69 & 16.03 \\
            ~& Transformer					&  7.29 & 12.75 & 17.33 & 12.35 &  9.63 & 14.83 & 19.43 & 14.53 \\
            ~& SR-LSTM \cite{zhang2019sr} 	&  9.50 & 15.67 & 22.48 & 15.76 & 11.56 & 18.16 & 21.82 & 17.05 \\
            ~& STAR \cite{yu2020spatio}		&  9.25 & 15.34 & 22.34 & 15.52 & 11.76 & 18.93 & 23.70 & 17.99 \\
            ~& EF-Transformer				&  \textbf{6.28} & \textbf{9.99} & \textbf{12.11} & \textbf{9.39}  & \textbf{7.42} & \textbf{10.83} & \textbf{12.56}  & \textbf{10.20} \\ 
            \hline
        \end{tabular*} 
    \end{center}
\end{table}

Table \ref{tab_dance_appendix} shows the comparisons between the task of trajectory inference and prediction on the dance dataset. It can be observed that performances of inference and prediction are similar in the dance dataset, which is different from results in the tennis dataset. Behavior of all dancers in a group dance have pre-defined patterns. Therefore, the information of future frames of observed dancers are not as important as in tennis dataset. 

\begin{table}[htb]
    \renewcommand\arraystretch{1.3}
    \setlength\tabcolsep{3pt}
    \caption{Comparisons with baselines and SOTA methods on pedestrian dataset.} 
    \label{tab_pedestrian_1frame}
    \begin{center}
        \scriptsize
        \begin{tabular*}{0.88\textwidth}{@{\extracolsep{\fill}}c | c  c  c  c  c  c  }
            \hline
            \multirow{2}*{\textbf{Methods}} & \multicolumn{6}{c}{\textbf{Performance MAD/FAD}} \\
            \cline{2-7}
            ~ & ETH & HOTEL & ZARA & ZARA2 & UNIV & AVG \\ 
            \hline\hline             
            SR-LSTM \cite{zhang2019sr} & 2.98/5.04&2.68/\textbf{4.70}&2.01/3.34&1.84/3.48&2.13/3.85&2.33/4.10 \\
            STAR \cite{yu2020spatio}	& 4.04/6.12&3.87/5.99&4.69/8.44&4.05/7.11&4.81/9.02&4.29/7.34 \\
            Transformer				& 2.52/4.54&2.30/3.76&1.75/3.01&1.85/3.55&2.23/4.35&2.13/3.84 \\
            EF-Transformer				& \textbf{2.30}/\textbf{4.32}&\textbf{2.66}/4.71&\textbf{1.50}/\textbf{2.58}&\textbf{1.48}/\textbf{2.75}&\textbf{2.07}/\textbf{3.63}&\textbf{2.00}/\textbf{3.60} \\
            
            \hline
        \end{tabular*} 
    \end{center}
\end{table}

Table \ref{tab_pedestrian_1frame} shows the trajectory prediction results on the pedestrian dataset, where only the first frame of ground truth is provided for target participants. Although our EF-Transformer outperformed all compared methods except FAD on `hotel', both MAD and FAD for all methods are significantly larger than results with 8-frame setting. As walking pattern of pedestrians highly relies on self intention information, which underlies the historical trajectories, it is difficult to do the prediction without history information. It is also can be observed in Fig.\ \ref{fig_pedvis} that if observed pedestrians are irrelevant to the target pedestrian, results of all methods with 1-frame setting are likely to fail, \eg, image at row 2 column 4.

\subsection{Multi-Task Inference and Prediction}\label{sec:multitask}

\begin{table}[htb]
    \renewcommand\arraystretch{1.4}
    \setlength\tabcolsep{3pt}
    \caption{Comparisons of multi-task inference and prediction with typical transformer on tennis dataset.`Traj' represents the task of trajectory inference or prediction, during which ground truth action labels are provided. `Multi' represents the task of multi-task inference or prediction, where both trajectories and action labels have to be estimated.} 
    \label{tab_multitask_appendix}
    \begin{center}
        \scriptsize
        \begin{tabular*}{0.88\textwidth}{@{\extracolsep{\fill}}c | c | c  c  c  c | c  c  c  c  }
            \hline
            \multirow{2}*{\rotatebox{90}{\textbf{ }}} & 	\multirow{2}*{\textbf{Inference}}&  \multicolumn{4}{c|}{\textbf{MAD}} & \multicolumn{4}{c}{\textbf{FAD}} \\
            \cline{3-10}
            ~ & ~ 	& \textbf{Short} & \textbf{Mid} & \textbf{Long} & \textbf{Avg} & \textbf{Short} & \textbf{Mid} & \textbf{Long} & \textbf{Avg} \\
            \hline
            \multirow{2}*{\rotatebox{90}{\textbf{Traj}}}
            & Transformer				& 21.17 & 32.91 & 46.67 & 30.95 & 37.14 & 52.06 & 68.14 & 49.34 \\
            ~& EF-Transformer			& 19.40 & \textbf{30.04} & 43.04 & \textbf{28.35} & 35.38 & \textbf{48.62} & 64.23 & \textbf{46.43} \\ 
            \hline
            \multirow{2}*{\rotatebox{90}{\textbf{Multi}}}
            & Transformer				& 20.22 & 33.42 & 49.76 & 31.36 & 36.11 & 52.65 & 72.79 & 50.02 \\
            ~& EF-Transformer			& \textbf{19.21} & 31.31 & \textbf{42.85} & 28.86 & \textbf{33.62} & 50.86 & \textbf{63.88} & 46.80 \\
            \hline
            \hline
            \multirow{2}*{\rotatebox{90}{\textbf{ }}} & 	\multirow{2}*{\textbf{Prediction}}&  \multicolumn{4}{c|}{\textbf{MAD}} & \multicolumn{4}{c}{\textbf{FAD}} \\
            \cline{3-10}
            ~ & ~ 	& \textbf{Short} & \textbf{Mid} & \textbf{Long} & \textbf{Avg} & \textbf{Short} & \textbf{Mid} & \textbf{Long} & \textbf{Avg} \\
            \hline
            \multirow{2}*{\rotatebox{90}{\textbf{Traj}}}
            & Transformer				& 20.14 & 33.09 & 50.70 & 31.33 & 35.85 & 52.55 & 71.57 & 49.67 \\
            ~& EF-Transformer			& \textbf{19.24} & 30.71 & \textbf{41.98} & \textbf{28.44} & 34.97 & 50.36 & \textbf{62.60} & 46.83 \\ 
            \hline
            \multirow{2}*{\rotatebox{90}{\textbf{Multi}}}
            & Transformer				& 20.32 & 33.71 & 50.81 & 31.71 & 36.50 & 55.10 & 76.96 & 52.01 \\
            ~& EF-Transformer			& 19.26 & \textbf{30.14} & 43.83 & 28.49 & \textbf{33.27} & \textbf{48.30} & 63.82 & \textbf{45.45} \\
            \hline
        \end{tabular*} 
    \end{center}
\end{table}

\begin{figure}[t]
    \centering
    \subfigure[Transformer]{\label{fig_dance_cm_typical}\includegraphics[width=35mm]{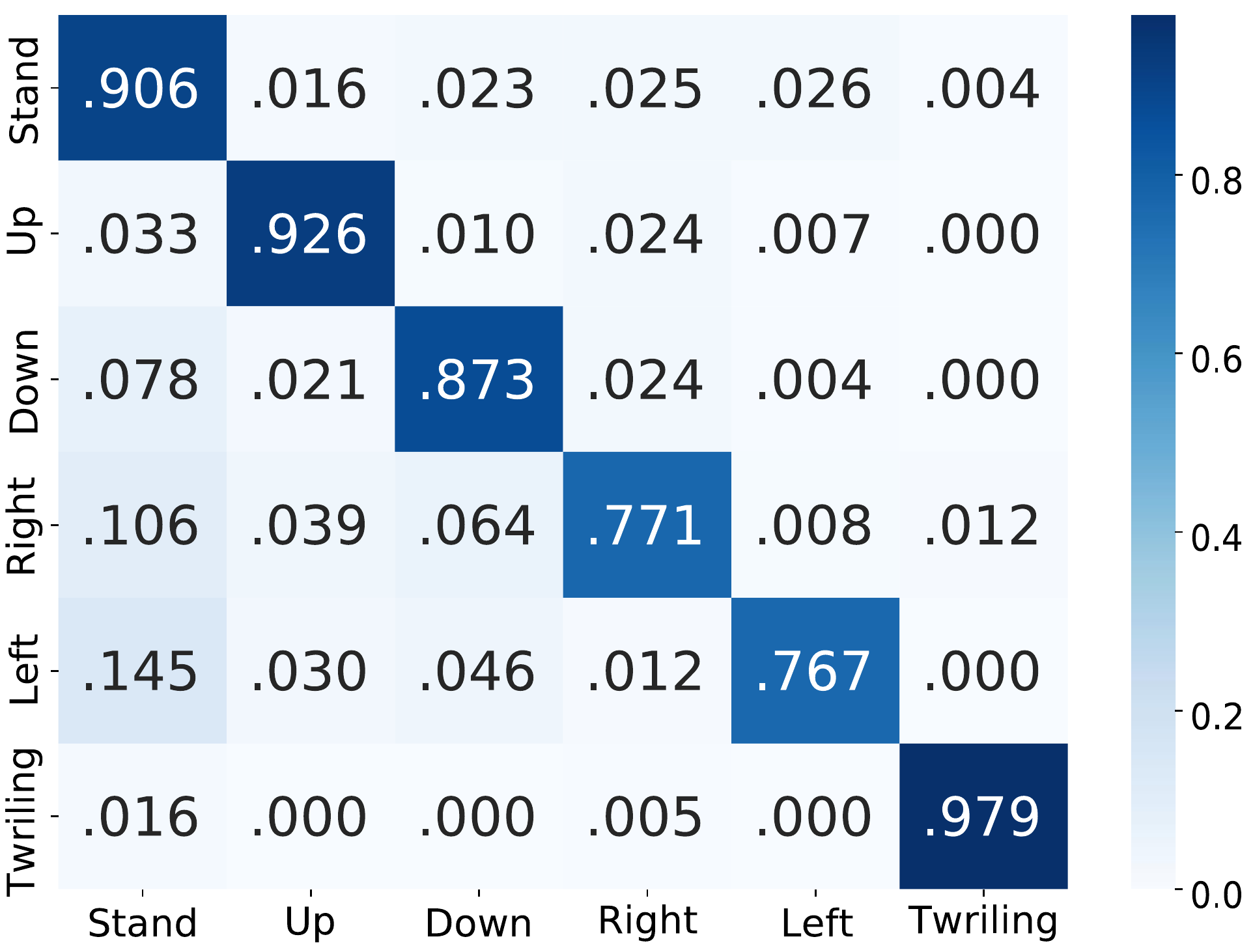}}\hspace{35pt}
    \subfigure[EF-Transformer]{\label{fig_dance_cm_ef}\includegraphics[width=35mm]{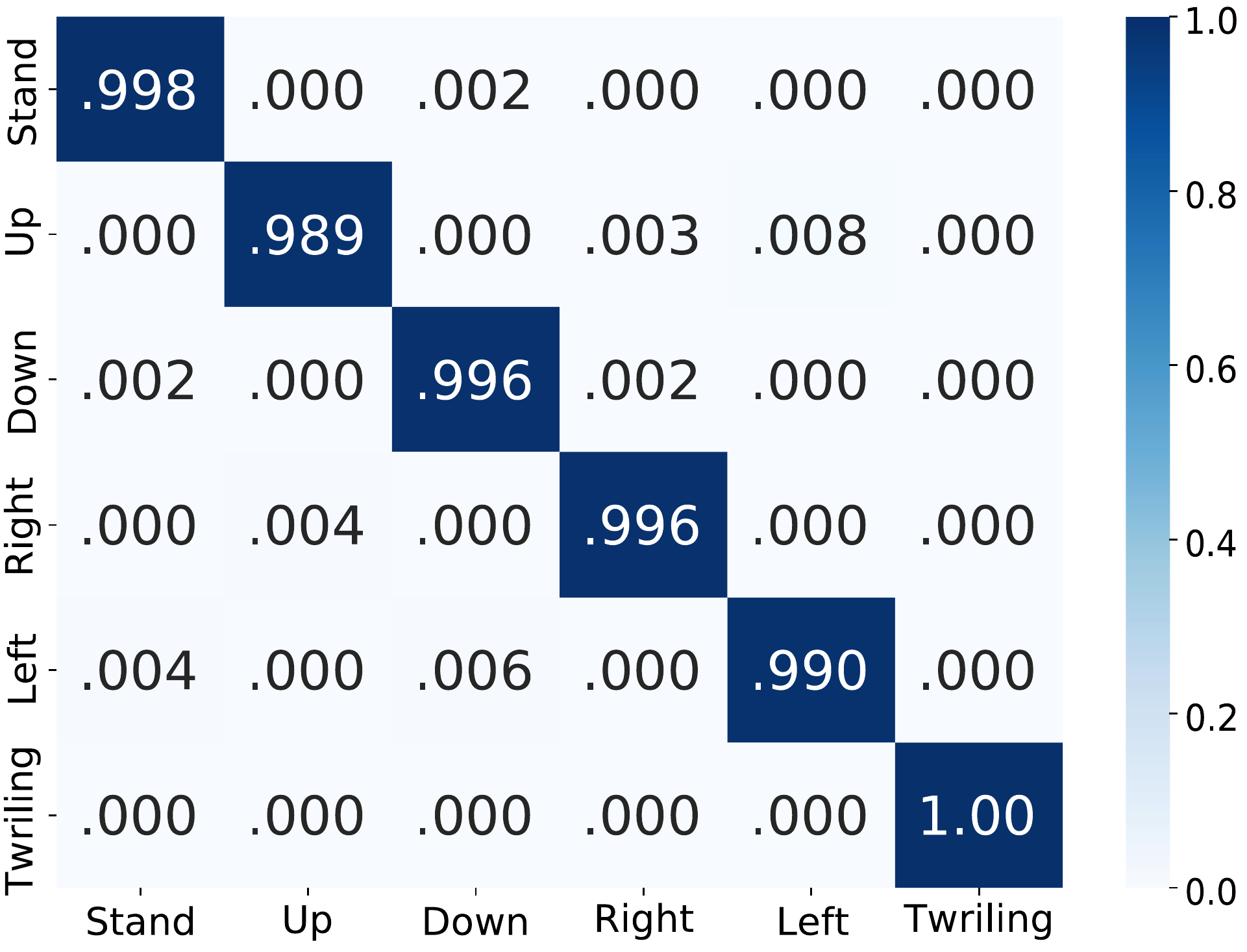}}
    \vspace{-10pt}
    \caption{Confusion matrices of action prediction on dance dataset. }
    \label{fig_dance_cm}
    \vspace{-15pt}
\end{figure}

The results of multi-task inference and prediction on the tennis dataset is shown in Table \ref{tab_multitask_appendix}, where a typical transformer is compared. As defined in Section \ref{sec:tennis}, action labels of different participants in one frame is complementary. For example in the serving stage, if the actions of observed participants are `S' and `WOS', then the action of the target participant will be `WTS'. The action of the target participant can easily be deduced from seeing the actions of other observed participants, so both the typical transformer and our EF-Transformer achieved 100\% accuracy for action inference and prediction. For our EF-Transformer, the trajectory inference and prediction results are also comparable between providing action ground truth and inferring actions simultaneously. The confusion matrices of two methods for action prediction is shown in Figure \ref{fig_dance_cm}.

\subsection{Robustness with Multiple Noise}
We follow the setting of Section 4.6 in our paper and evaluate the performances of our method and typical transformer with multiple-frame noise involved. The results are shown in Table \ref{tab_robust_multiple}.

\begin{table}[htb]
    \vspace{-10pt}
    \renewcommand\arraystretch{1.23}
    \setlength\tabcolsep{3pt}
    \caption{Comparisons of FAD on tennis dataset with multiple noise involved in different frames.} 
    \label{tab_robust_multiple}
    \begin{center}
        \scriptsize
        \begin{tabular*}{0.88\textwidth}{@{\extracolsep{\fill}} c | c  c  c  c | c  c  c  c  }
            \hline
            \multirow{2}*{\textbf{\makecell*[c]{Noise \\ Position}}}&  \multicolumn{4}{c|}{\textbf{Transformer FAD}} & \multicolumn{4}{c}{\textbf{EF-Transformer FAD}} \\
            \cline{2-9}
            ~ 	& \textbf{Short} & \textbf{Mid} & \textbf{Long} & \textbf{Avg} & \textbf{Short} & \textbf{Mid} & \textbf{Long} & \textbf{Avg} \\
            \hline\hline
            No Noise	& 37.14 & 52.06 & 68.14 & 49.34 & 35.38 & 48.62 & 64.23 & 46.43 \\
            Noise@t=2,3	& 95.66 &122.22 &162.25 &119.27 & 39.88 & 65.46 & 99.97 & 61.95 \\
            Noise@t=5,6	& 75.86 &101.57 &141.87 & 98.97 & 50.62 & 65.48 & 94.78 & 64.97 \\
            Noise@t=8,9	&134.23 &164.20 &211.25 &161.18 &132.16 &143.73 &170.83 &144.05 \\
            \hline
        \end{tabular*} 
    \end{center}
    \vspace{-20pt}
\end{table}

\section{Additional Visualizations}\label{sec:vis}
We also provide additional visualizations of results on different datasets.

\begin{figure*}[ht]
    \centering
    \begin{minipage}[t]{0.48\textwidth}
        \centering
        \includegraphics[width=5.5cm]{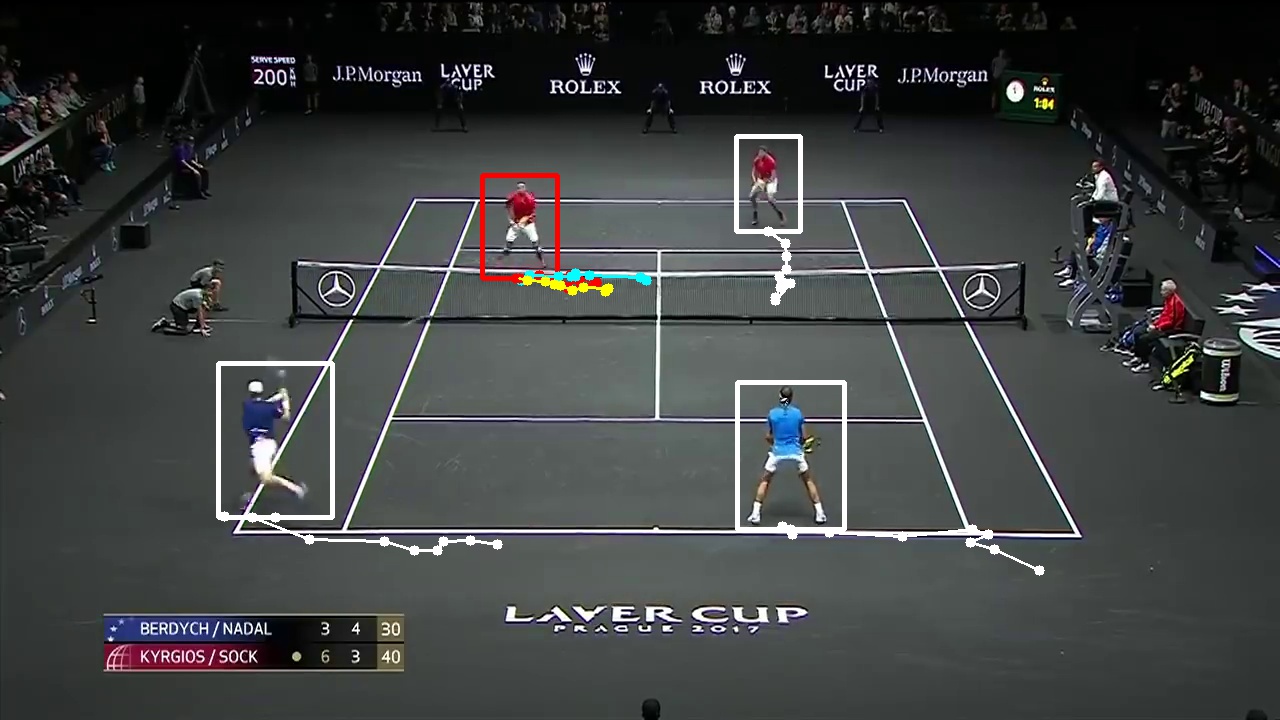}
        \centerline{}
    \end{minipage}
    \begin{minipage}[t]{0.48\textwidth}
        \centering
        \includegraphics[width=5.5cm]{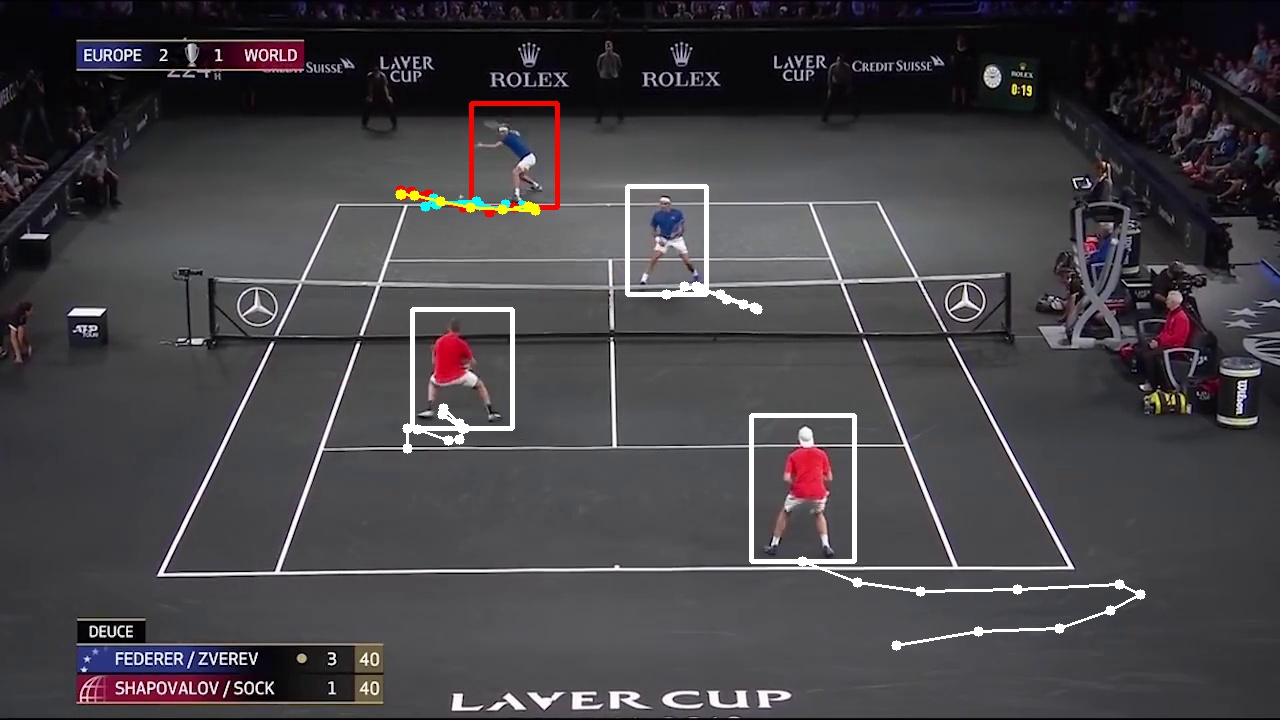}
        \centerline{} \\
    \end{minipage}
    \begin{minipage}[t]{0.48\textwidth}
        \centering
        \includegraphics[width=5.5cm]{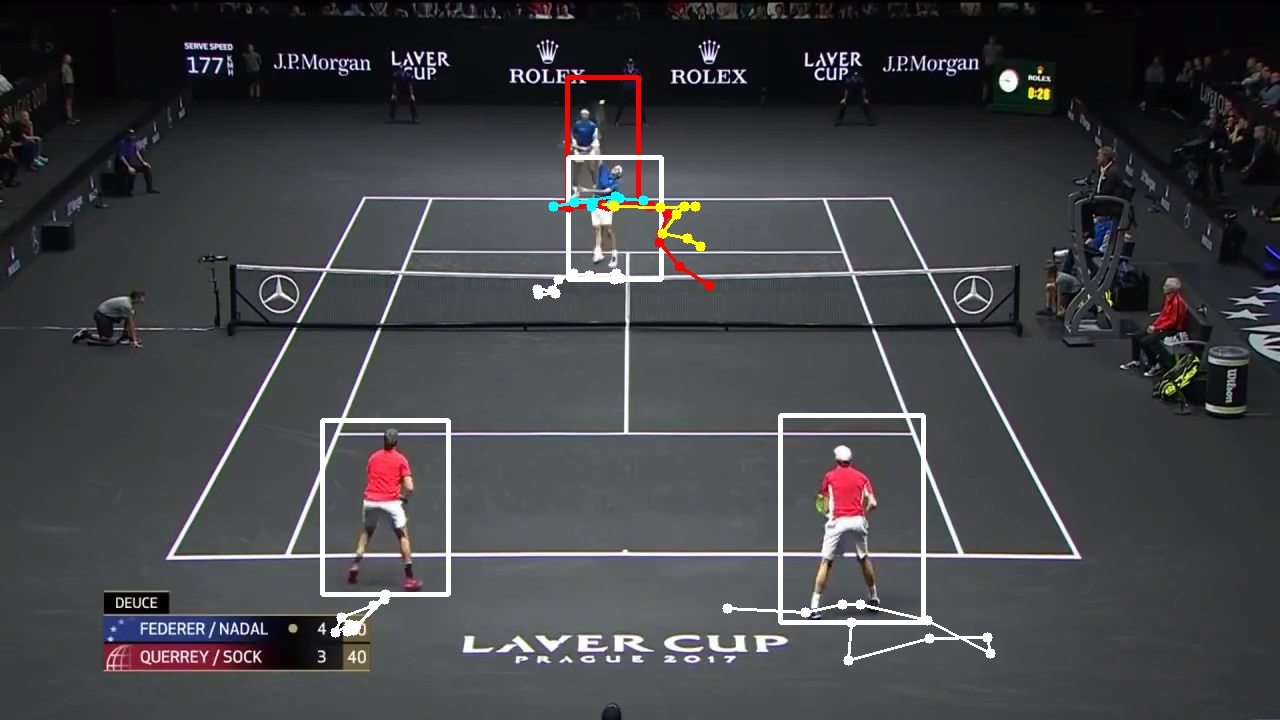}
        \centerline{}
    \end{minipage}
    \begin{minipage}[t]{0.48\textwidth}
        \centering
        \includegraphics[width=5.5cm]{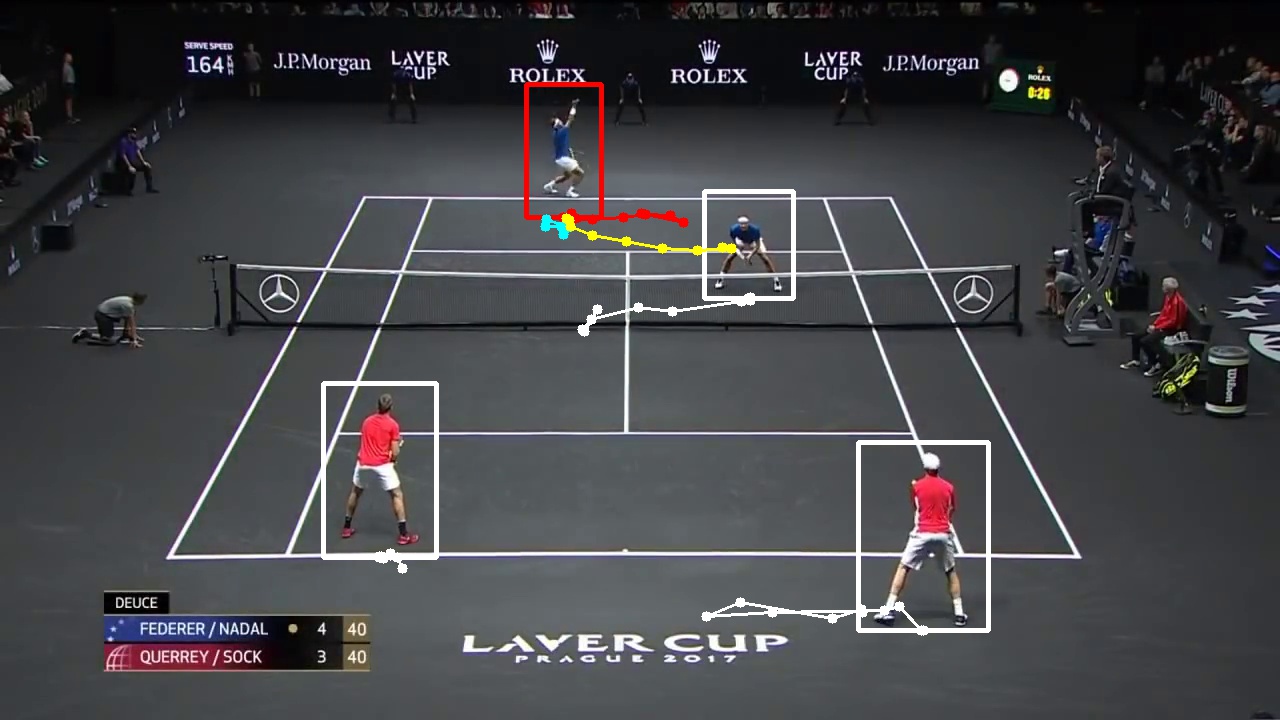}
        \centerline{} \\
    \end{minipage}
    \begin{minipage}[t]{0.48\textwidth}
        \centering
        \includegraphics[width=5.5cm]{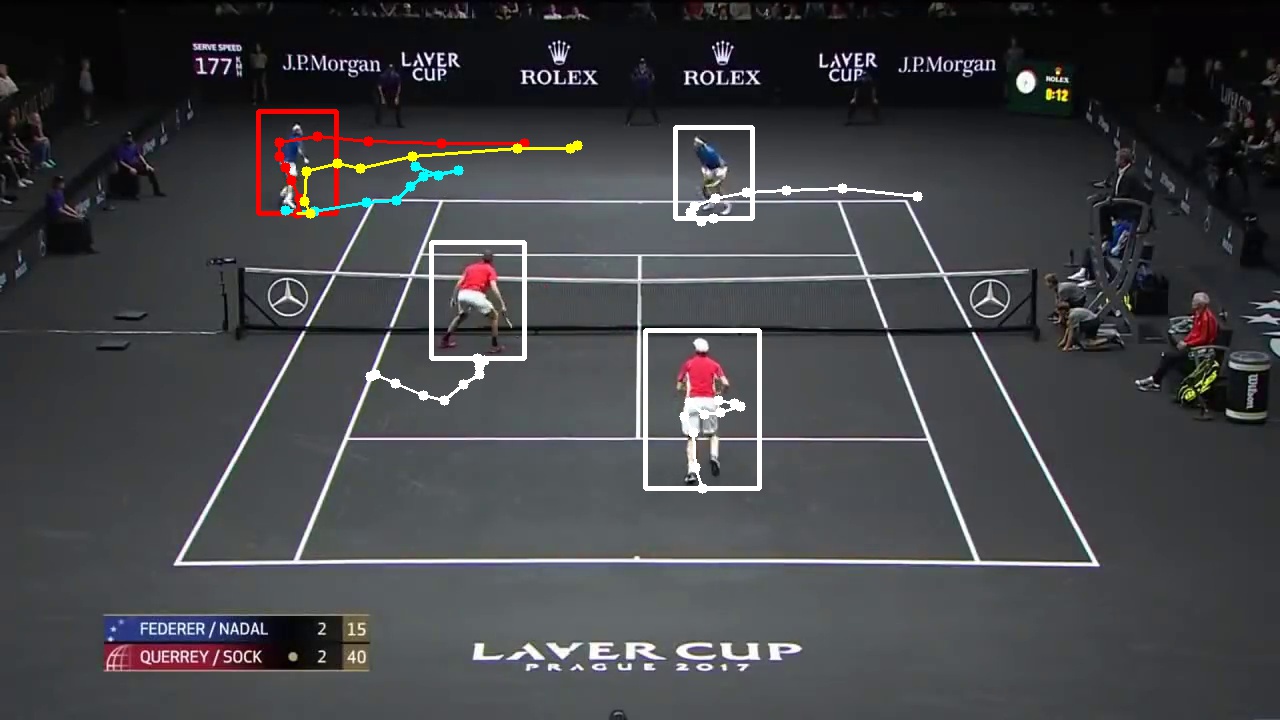}
        \centerline{}
    \end{minipage}
    \begin{minipage}[t]{0.48\textwidth}
        \centering
        \includegraphics[width=5.5cm]{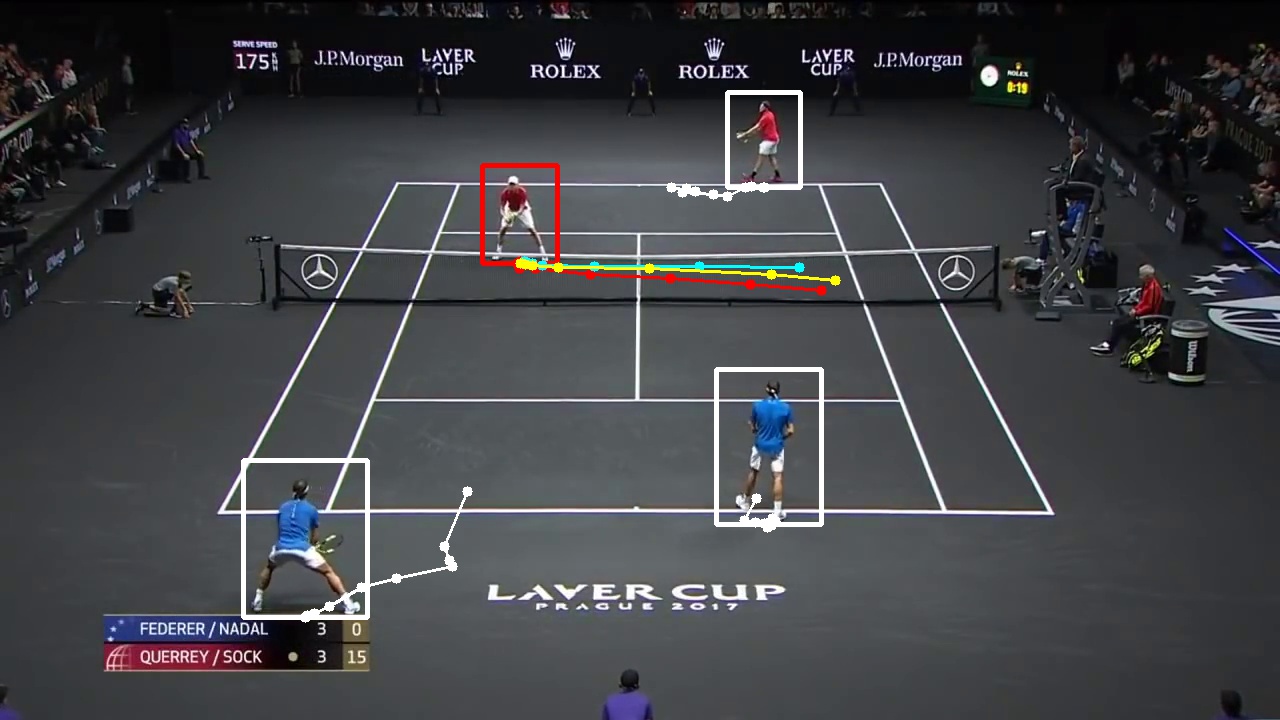}
        \centerline{} \\
    \end{minipage}
    \begin{minipage}[t]{0.48\textwidth}
        \centering
        \includegraphics[width=5.5cm]{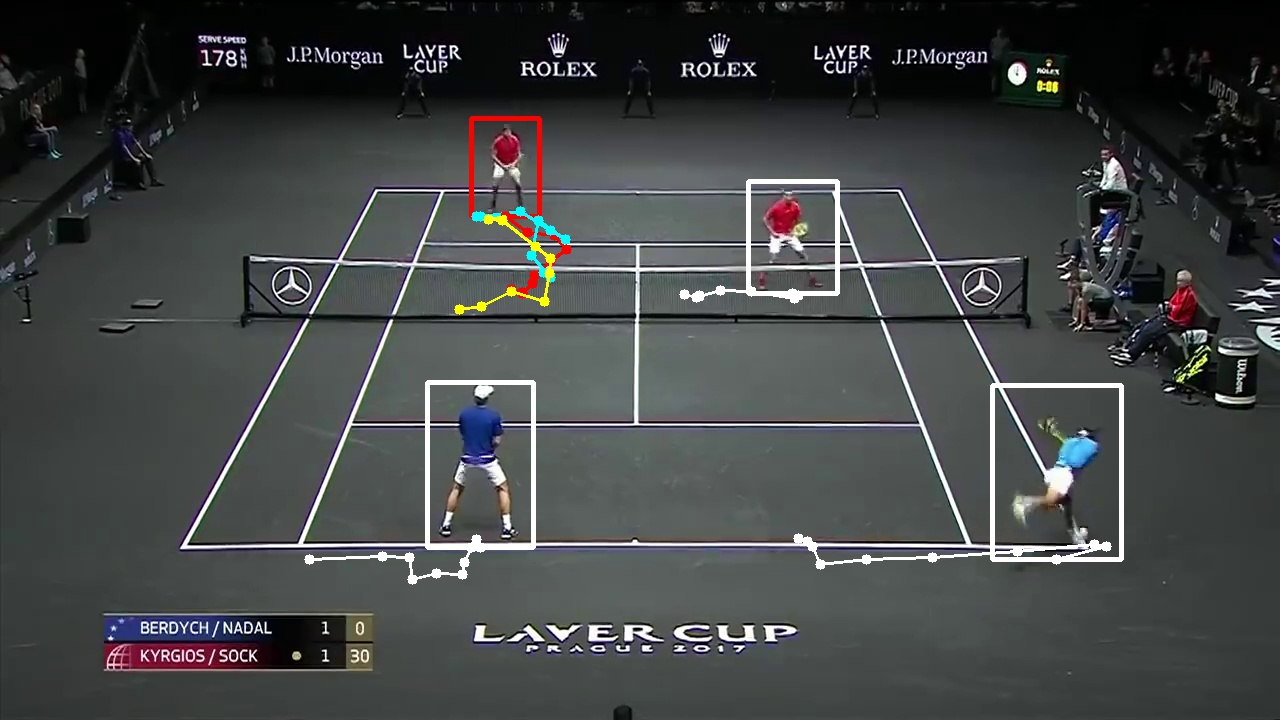}
        \centerline{}
    \end{minipage}
    \begin{minipage}[t]{0.48\textwidth}
        \centering
        \includegraphics[width=5.5cm]{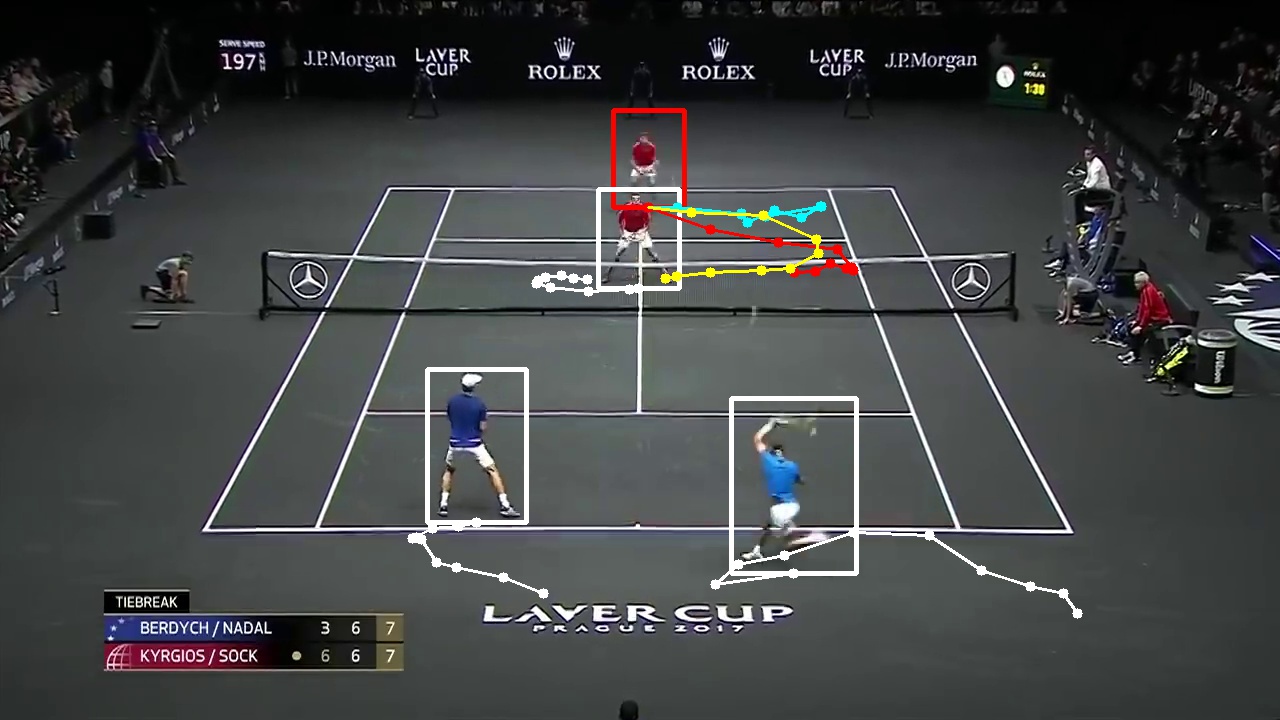}
        \centerline{} \\
    \end{minipage}
    \caption{Visualization of trajectory prediction results of EF-Transformer and typical transformer on tennis dataset. White rectangles and trajectories are the observed participants. Red rectangles are target participants with red trajectories for ground truth. Cyan trajectories are predicted by typical transformer and yellow ones are predicted by our method.}    \label{fig_tennisvis_appendix}
\end{figure*}

\begin{figure*}[ht]
    \centering
    \begin{minipage}[t]{0.48\textwidth}
        \centering
        \includegraphics[width=5.5cm]{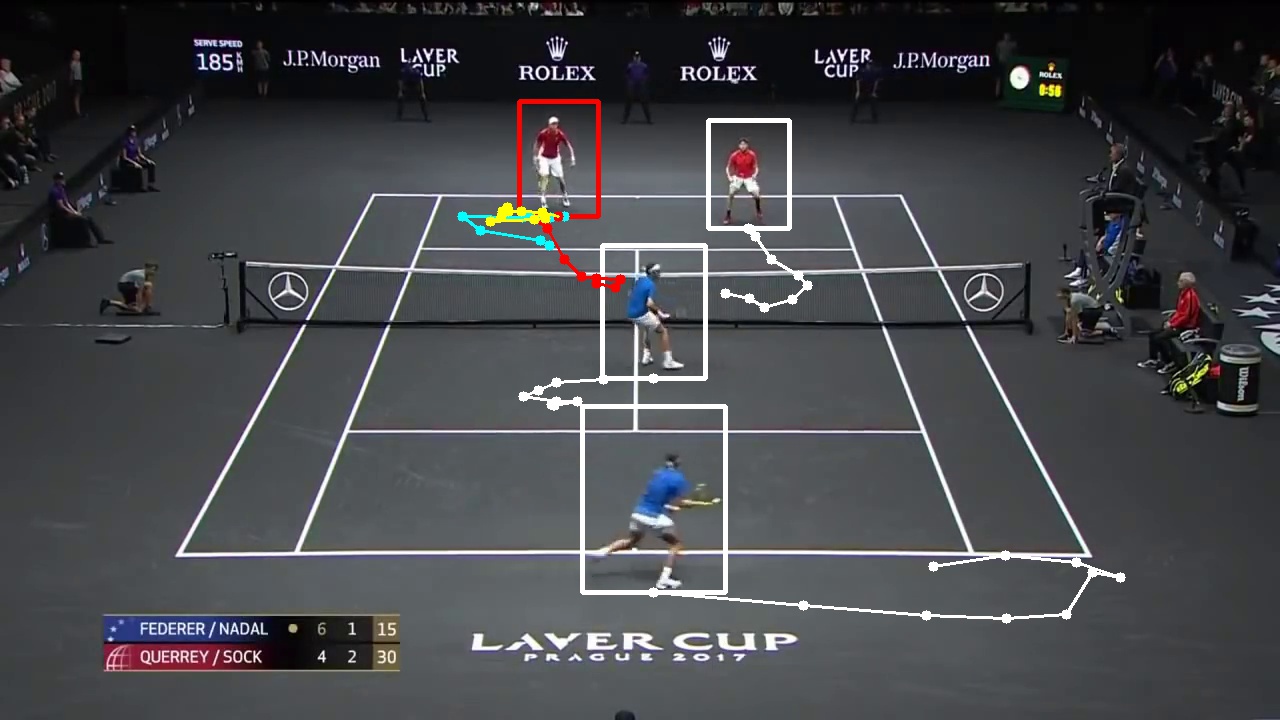}
        \centerline{}
    \end{minipage}
    \begin{minipage}[t]{0.48\textwidth}
        \centering
        \includegraphics[width=5.5cm]{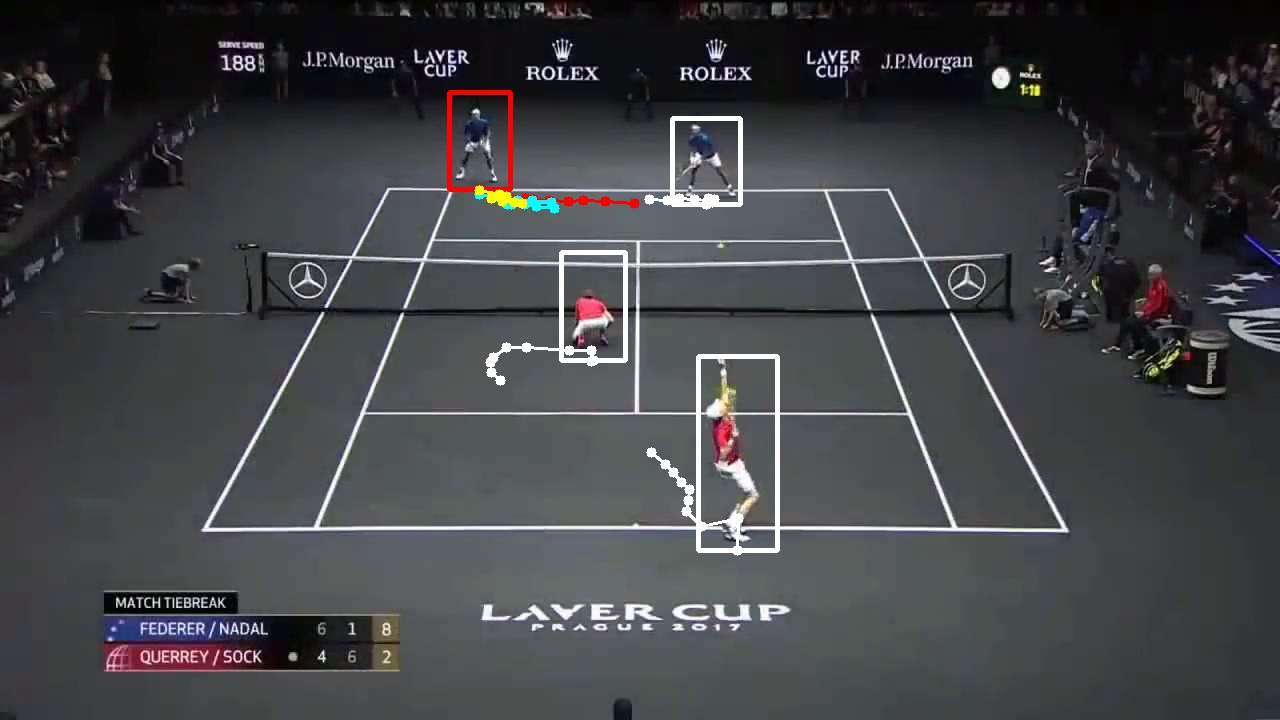}
        \centerline{} \\
    \end{minipage}
    \begin{minipage}[t]{0.48\textwidth}
        \centering
        \includegraphics[width=5.5cm]{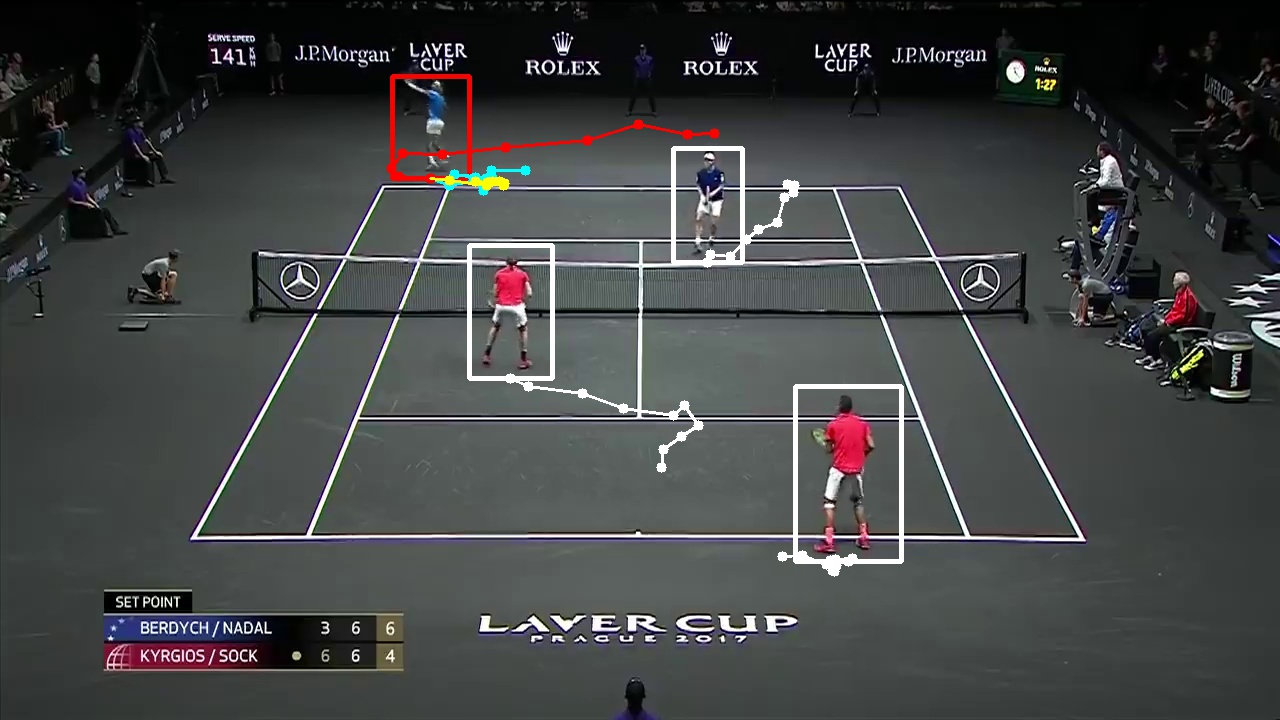}
        \centerline{}
    \end{minipage}
    \begin{minipage}[t]{0.48\textwidth}
        \centering
        \includegraphics[width=5.5cm]{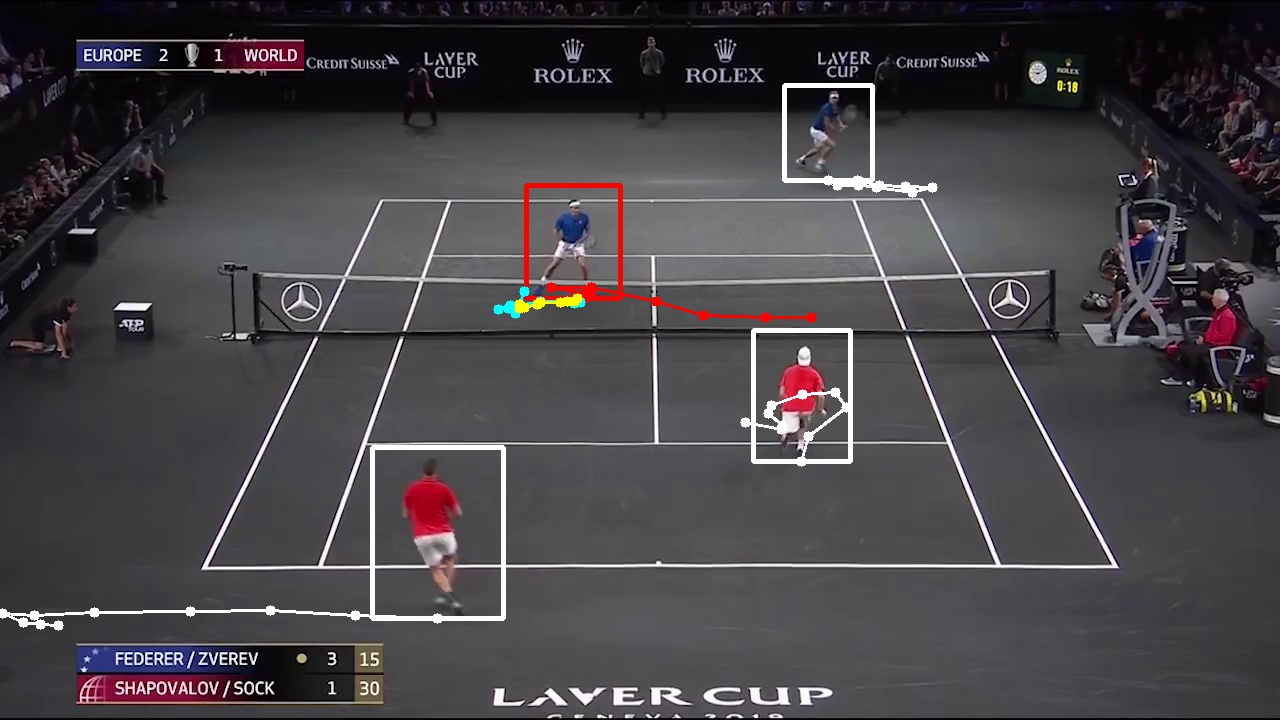}
        \centerline{} \\
    \end{minipage}
    \caption{Some failure cases of trajectory prediction in tennis dataset. White rectangles and trajectories are the observed participants. Red rectangles are target participants with red trajectories for ground truth. Cyan trajectories are predicted by typical transformer and yellow ones are predicted by our method. 
    \protect\\The top two figures show that our model fails when the target player go towards his teammate as usually two players try to defense as much area as possible instead of stand close to each other. 
    \protect\\The bottom left figure shows that our model suppose that the teammate of the target player will pursue the ball but in fact both of players try to get the ball in this round.
    \protect\\The bottom right figure shows that our model choose to play safe and let his teammates to hit the ball back, however, the real player decide to intercept the ball close to the net to win this round.}
    \label{fig_tennisfail}
\end{figure*}

Fig.\ \ref{fig_tennisvis_appendix} shows some trajectory predictions on tennis dataset. Compared to typical transformer, the advantages of our EF-Transformer is significant especially for long and not smooth trajectories (rows 3 and 4 in Fig.\ \ref{fig_tennisvis_appendix}). Some failure cases are shown and discussed in Fig.\ \ref{fig_tennisfail} Some videos are also provided to dynamically illustrate the trajectory prediction. In the videos, boxes show the positions of participants in the current frame, for which the red box is the target participant and white boxes are observed participants. Similarly, the red trajectory is the ground truth for the target participant and the white trajectories are for the observed participants. The trajectory predicted by our EF-Transformer is in green, and the prediction of the current frame is represented by a yellow arrow. The trajectory predicted by the typical transformer is in blue, and the prediction of the current frame is a cyan arrow. Note that when predicting the trajectory of the target participant in the current frame, the provided input information comprises white observed trajectories, the predicted target trajectory in past frames (blue or green), the coarse ball trajectory, and action labels, while the red ground truth target trajectory is hidden and only shown for comparison.

Some results of NBA dataset are visualized in videos, where we can see that both of our EF-Transformer and typical transformer achieve good prediction performance. Some failure cases happen when the target player go to defend another opponent, which sometimes is also reasonable in real basketball games.

\begin{figure*}[ht]
    \centering
    \begin{minipage}[t]{0.32\textwidth}
        \centering
        \includegraphics[width=3.5cm]{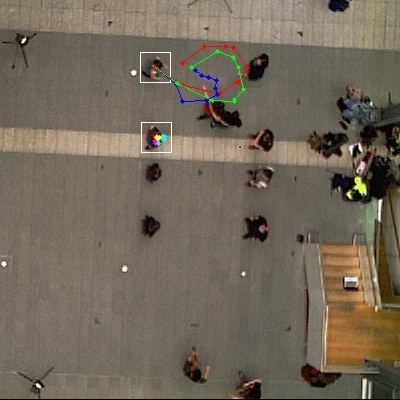}
        \centerline{}
    \end{minipage}
    \begin{minipage}[t]{0.32\textwidth}
        \centering
        \includegraphics[width=3.5cm]{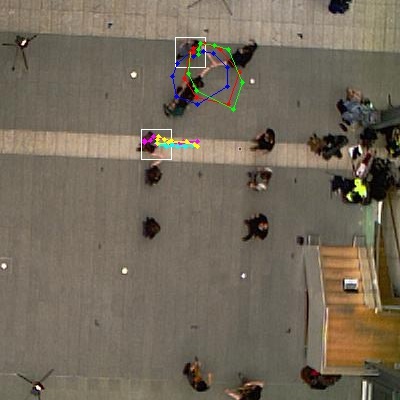}
        \centerline{} 
    \end{minipage}
    \begin{minipage}[t]{0.32\textwidth}
        \centering
        \includegraphics[width=3.5cm]{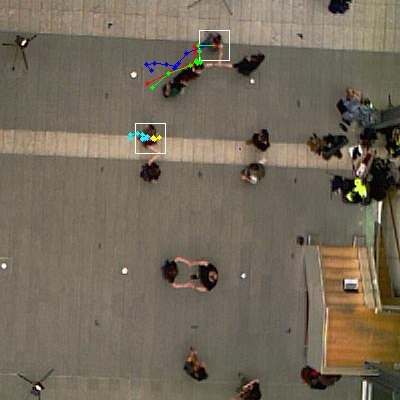}
        \centerline{} \\
    \end{minipage}
    \begin{minipage}[t]{0.32\textwidth}
        \centering
        \includegraphics[width=3.5cm]{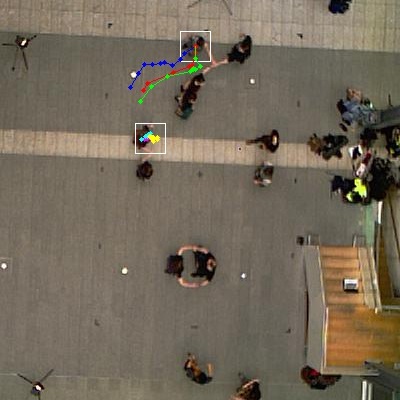}
        \centerline{} 
    \end{minipage}
    \begin{minipage}[t]{0.32\textwidth}
        \centering
        \includegraphics[width=3.5cm]{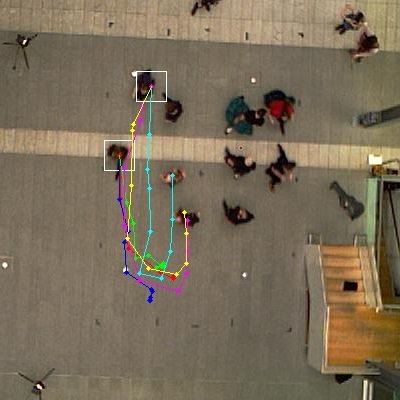}
        \centerline{}
    \end{minipage}
    \begin{minipage}[t]{0.32\textwidth}
        \centering
        \includegraphics[width=3.5cm]{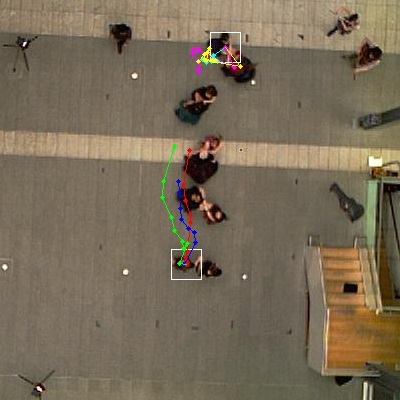}
        \centerline{} \\
    \end{minipage}
    \begin{minipage}[t]{0.32\textwidth}
        \centering
        \includegraphics[width=3.5cm]{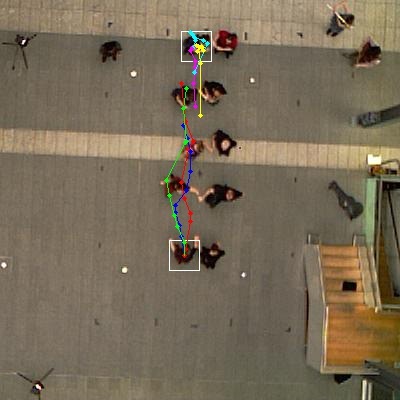}
        \centerline{}
    \end{minipage}
    \begin{minipage}[t]{0.32\textwidth}
        \centering
        \includegraphics[width=3.5cm]{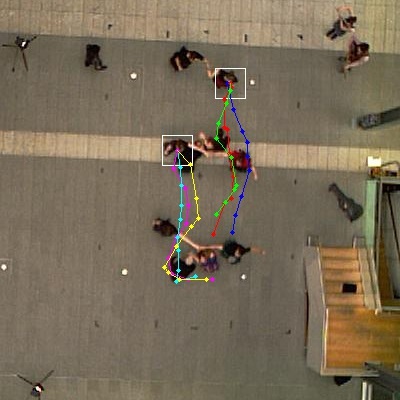}
        \centerline{} 
    \end{minipage}
    \begin{minipage}[t]{0.32\textwidth}
        \centering
        \includegraphics[width=3.5cm]{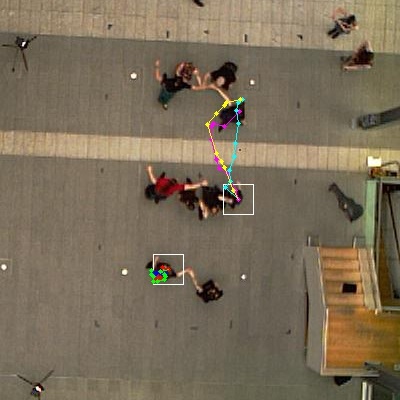}
        \centerline{} \\
    \end{minipage}
    \begin{minipage}[t]{0.32\textwidth}
        \centering
        \includegraphics[width=3.5cm]{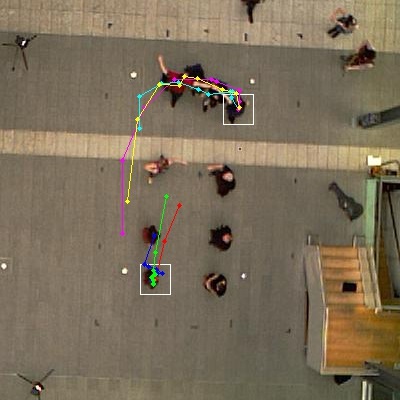}
        \centerline{} 
    \end{minipage}
    \begin{minipage}[t]{0.32\textwidth}
        \centering
        \includegraphics[width=3.5cm]{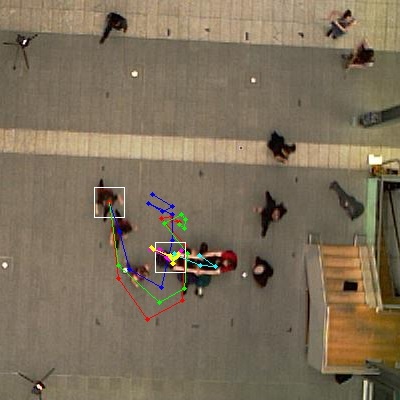}
        \centerline{}
    \end{minipage}
    \begin{minipage}[t]{0.32\textwidth}
        \centering
        \includegraphics[width=3.5cm]{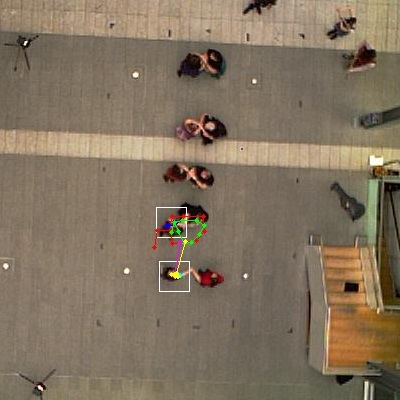}
        \centerline{} \\
    \end{minipage}
    \caption{Visualization of trajectory prediction results on dance datasets. White rectangles are initial positions of two target dancers. Ground truth trajectories are represented in red and magenta. Trajectories predicted by typical transformer are in blue and cyan, while by our EF-Transformer are in green and yellow. Frames are cropped to $400$$\times$$400$ for better view. }    \label{fig_dancevis}
\end{figure*}

Results of trajectory prediction of two target dancers on dance dataset are shown in Fig.\ \ref{fig_dancevis}, from which we can observe that our method is capable of predicting more precise moving trends and patterns than the typical transformer.

\begin{figure*}[ht]
    \centering
    \begin{minipage}[t]{0.24\textwidth}
        \centering
        \includegraphics[width=3cm]{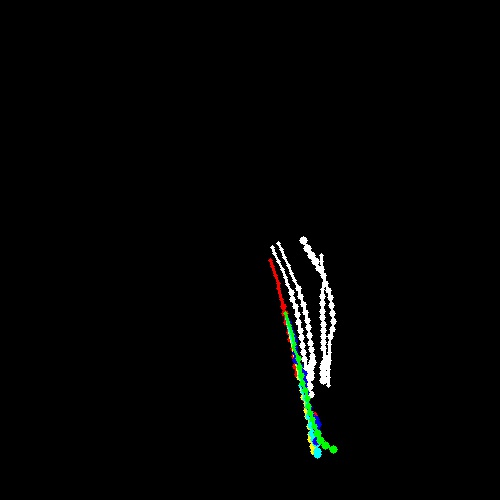}
        \centerline{}
    \end{minipage}
    \begin{minipage}[t]{0.24\textwidth}
        \centering
        \includegraphics[width=3cm]{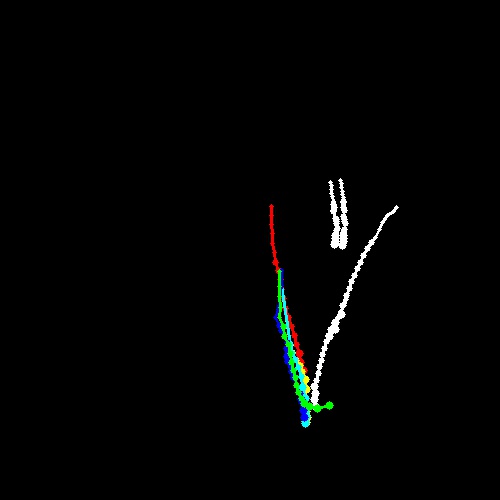}
        \centerline{} 
    \end{minipage}
    \begin{minipage}[t]{0.24\textwidth}
        \centering
        \includegraphics[width=3cm]{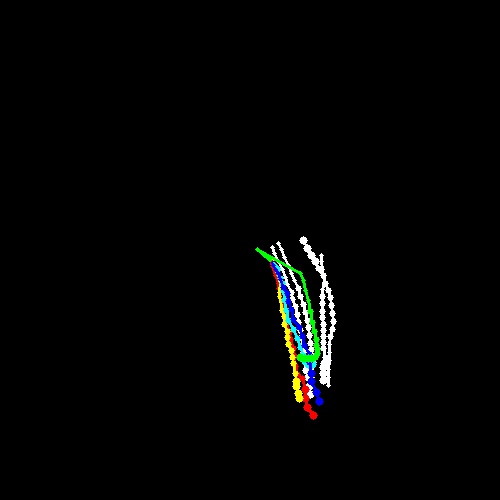}
        \centerline{}
    \end{minipage}
    \begin{minipage}[t]{0.24\textwidth}
        \centering
        \includegraphics[width=3cm]{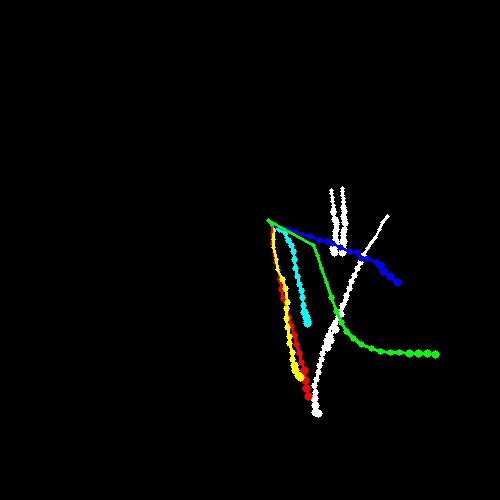}
        \centerline{} \\ 
    \end{minipage}
    \vspace{-5pt}
    \begin{minipage}[t]{0.24\textwidth}
        \centering
        \includegraphics[width=3cm]{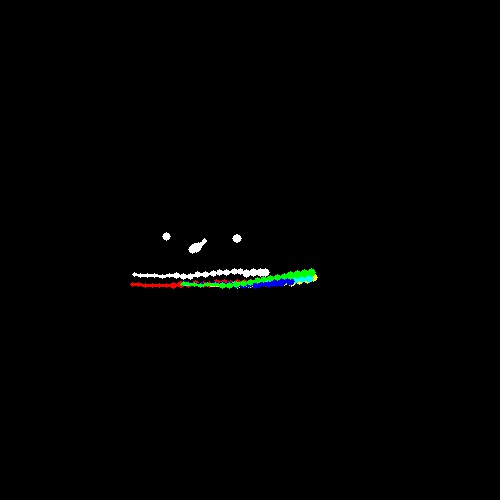}
        \centerline{}
    \end{minipage}
    \begin{minipage}[t]{0.24\textwidth}
        \centering
        \includegraphics[width=3cm]{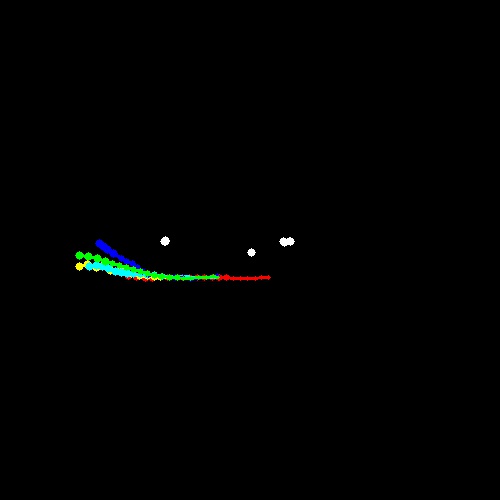}
        \centerline{} 
    \end{minipage}
    \begin{minipage}[t]{0.24\textwidth}
        \centering
        \includegraphics[width=3cm]{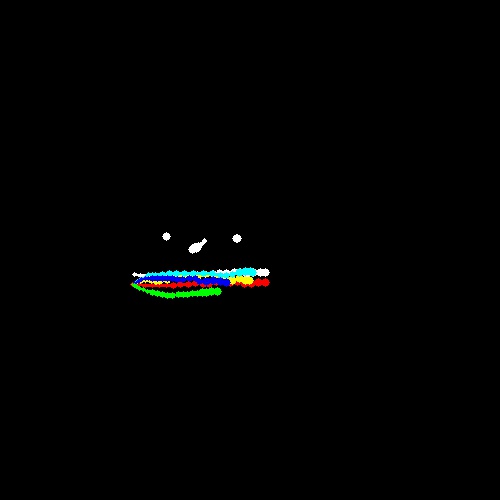}
        \centerline{}
    \end{minipage}
    \begin{minipage}[t]{0.24\textwidth}
        \centering
        \includegraphics[width=3cm]{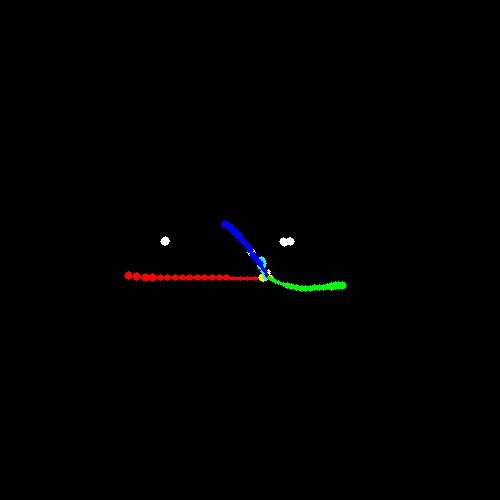}
        \centerline{} \\
    \end{minipage}
    \vspace{-5pt}
    \begin{minipage}[t]{0.24\textwidth}
        \centering
        \includegraphics[width=3cm]{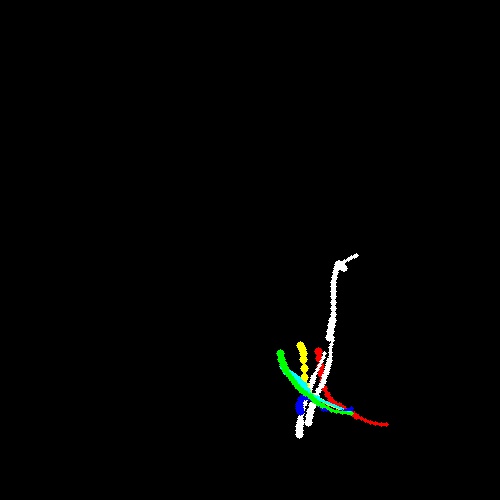}
        \centerline{}
    \end{minipage}
    \begin{minipage}[t]{0.24\textwidth}
        \centering
        \includegraphics[width=3cm]{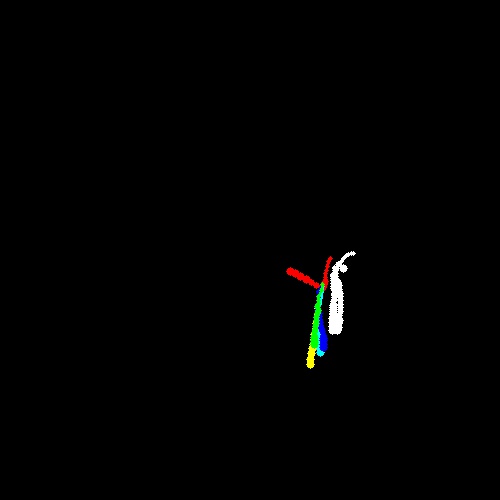}
        \centerline{} 
    \end{minipage}
    \begin{minipage}[t]{0.24\textwidth}
        \centering
        \includegraphics[width=3cm]{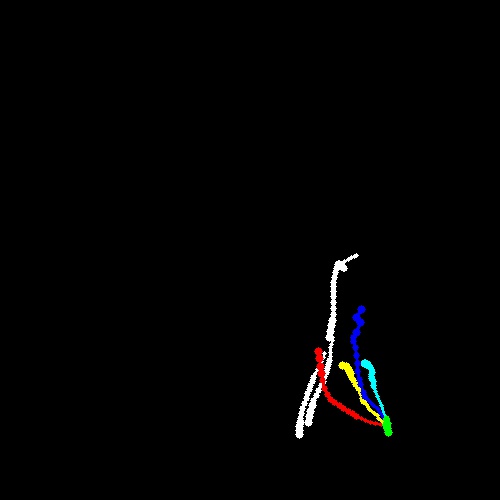}
        \centerline{}
    \end{minipage}
    \begin{minipage}[t]{0.24\textwidth}
        \centering
        \includegraphics[width=3cm]{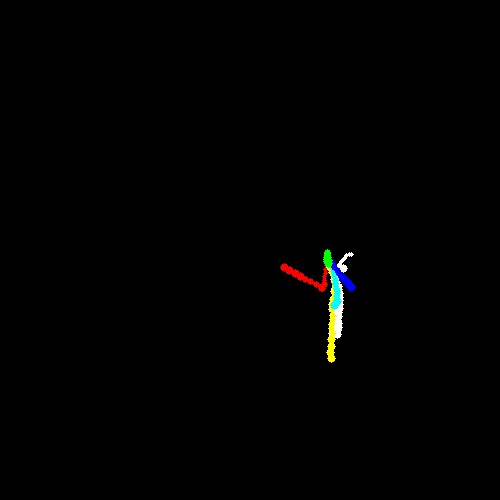}
        \centerline{} \\
    \end{minipage}
    \vspace{-5pt}
    \begin{minipage}[t]{0.24\textwidth}
        \centering
        \includegraphics[width=3cm]{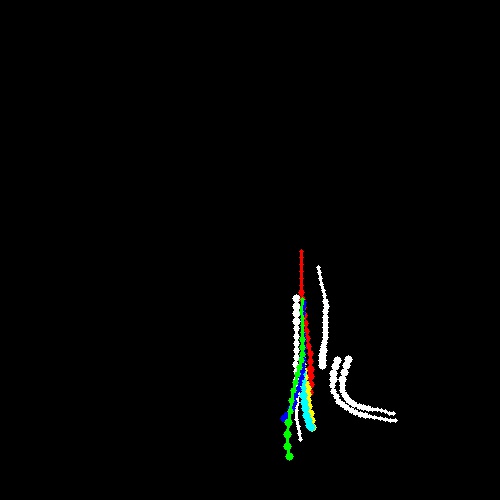}
        \centerline{}
    \end{minipage}
    \begin{minipage}[t]{0.24\textwidth}
        \centering
        \includegraphics[width=3cm]{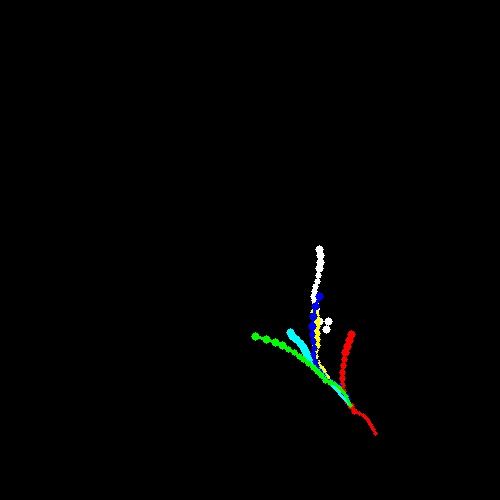}
        \centerline{} 
    \end{minipage}
    \begin{minipage}[t]{0.24\textwidth}
        \centering
        \includegraphics[width=3cm]{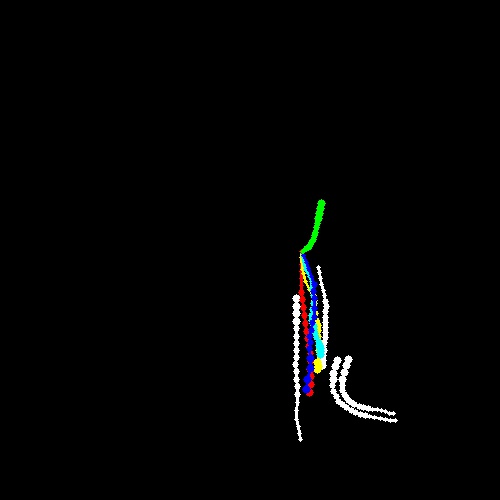}
        \centerline{}
    \end{minipage}
    \begin{minipage}[t]{0.24\textwidth}
        \centering
        \includegraphics[width=3cm]{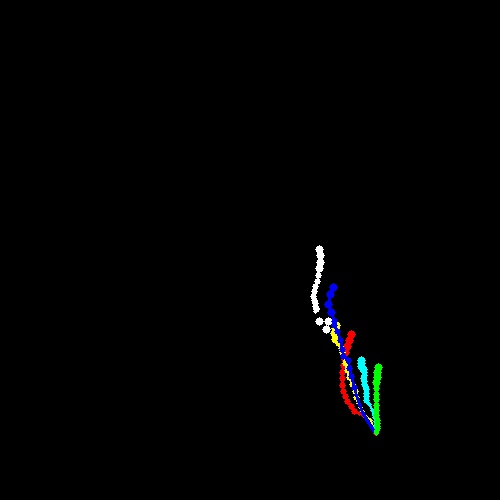}
        \centerline{} \\
    \end{minipage}
    \vspace{-5pt}
    \begin{minipage}[t]{0.24\textwidth}
        \centering
        \includegraphics[width=3cm]{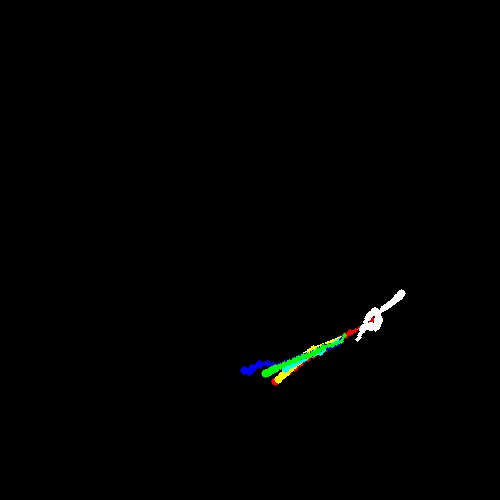}
        \centerline{}
    \end{minipage}
    \begin{minipage}[t]{0.24\textwidth}
        \centering
        \includegraphics[width=3cm]{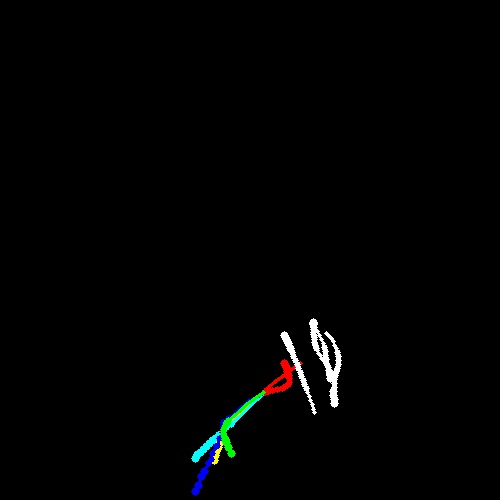}
        \centerline{} 
    \end{minipage}
    \begin{minipage}[t]{0.24\textwidth}
        \centering
        \includegraphics[width=3cm]{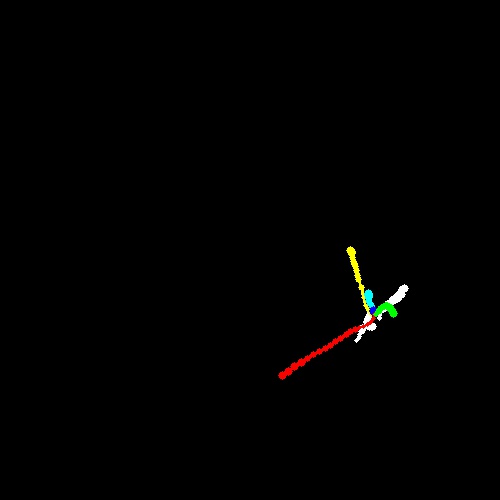}
        \centerline{}
    \end{minipage}
    \begin{minipage}[t]{0.24\textwidth}
        \centering
        \includegraphics[width=3cm]{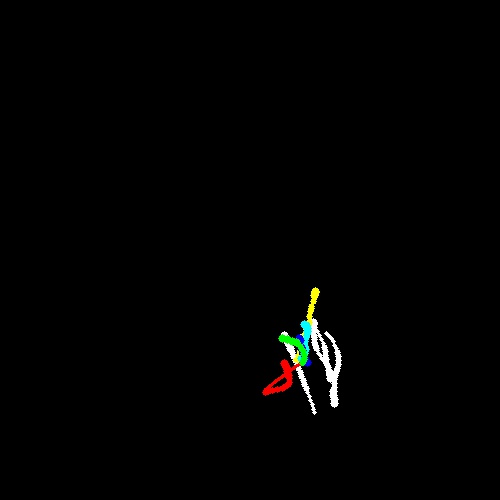}
        \centerline{} \\
    \end{minipage}
    \vspace{-5pt}
    \caption{Visualization of trajectory prediction results on pedestrian datasets. White trajectories are observed pedestrians, and red ones represent the ground truth of the target pedestrian. Yellow, cyan, blue, and green trajectories are predicted by our EF-Transformer, typical transformer, SR-LSTM \cite{zhang2019sr} and STAR \cite{yu2020spatio} correspondingly. The first two columns are results with 8-frame setting and last two columns are the results of same samples with 1-frame setting.}    \label{fig_pedvis}
\end{figure*}

Fig.\ \ref{fig_pedvis} shows some results of trajectory prediction on pedestrian datasets, where each row of images are the examples of subset `eth', `hotel', `zara1', `zara2', and `univ' respectively. The first two columns are results with the 8-frame setting, \ie, 8-frame ground truth trajectory is provided for the target pedestrian. The last two columns are the results of same testing samples with the 1-frame setting. Qualitatively, our EF-Transformer provided best predictions on the 8-frame setting. For the 1-frame setting, our method can also predict the trajectory better than other methods when the selected observed pedestrians are relevant to the target pedestrian.

\end{document}